\documentclass{article}

\usepackage{arxiv}

\usepackage[utf8]{inputenc} 
\usepackage[T1]{fontenc}    
\usepackage{hyperref}       
\usepackage{url}            
\usepackage{booktabs}       
\usepackage{amsfonts}       
\usepackage{nicefrac}       
\usepackage{microtype}      
\usepackage{lipsum}		
\usepackage{graphicx}
\usepackage{natbib}
\usepackage{doi}

\usepackage{booktabs} 

\usepackage{tabularx} 

\usepackage{CJKutf8}
\usepackage{arydshln}

\title{RedWhale: An Adapted Korean LLM Through Efficient Continual Pretraining}

\author{Anh-Dung Vo, Minseong Jung, Wonbeen Lee \\
	DocumentAI Team\\
	AGILESODA INC.\\
	Seoul, 06149, South Korea \\
	\texttt{contact@agilesoda.ai} \\
	\AND
	Daewoo Choi \\
	Department of Statistics\\
	Hankuk University of Foreign Studies\\
	Seoul, 02450, South Korea \\
	\texttt{daewoo.choi@agilesoda.ai-hufs.ac.kr} \\
}

\renewcommand{\headeright}{Technical Report}
\renewcommand{\undertitle}{Technical Report}

\hypersetup{
pdftitle={RedWhale: An Adapted Korean LLM Through Efficient Continual Pretraining},
pdfsubject={LLM, KOREAN},
pdfauthor={AGILESODA, DOCUMENTAI},
pdfkeywords={LLM, KOREAN, PRETRAINING, EFFICIENT},
}

\begin{document}
\maketitle

\begin{abstract}
The field of Natural Language Processing (NLP) has seen significant advancements with the development of Large Language Models (LLMs). However, much of this research remains focused on English, often overlooking low-resource languages like Korean. This oversight presents challenges due to the unique non-alphabetic token structure of Korean and the substantial memory and computational demands required for LLM training, which frequently lead to memory constraints and out-of-memory errors. To address these issues, we present RedWhale, a model specifically tailored for Korean language processing. RedWhale is developed using an efficient continual pretraining approach that includes a comprehensive Korean corpus preprocessing pipeline, a specialized tokenizer, an optimized model initialization technique, and a multistage pretraining strategy. These innovations collectively reduce training time and computational costs while maintaining high levels of accuracy and comprehension. By leveraging cross-lingual transfer learning, which exploits shared linguistic similarities across languages, RedWhale builds on English models to enhance Korean language processing. Experimental results demonstrate that RedWhale outperforms other leading models on Korean NLP benchmarks, including the Korean Balanced Evaluation of Significant Tasks (KoBEST), showing superior understanding and generation of Korean text. Furthermore, RedWhale showed no signs of convergence even after pretraining on 9.7 billion tokens, indicating the potential for further improvements with additional training. This work represents a significant advancement in bridging the linguistic divide, particularly in enhancing NLP capabilities for the Korean language.
\end{abstract}


\section{Introduction}

\begin{figure}[ht]
\centering
\includegraphics[width=1.00\linewidth]{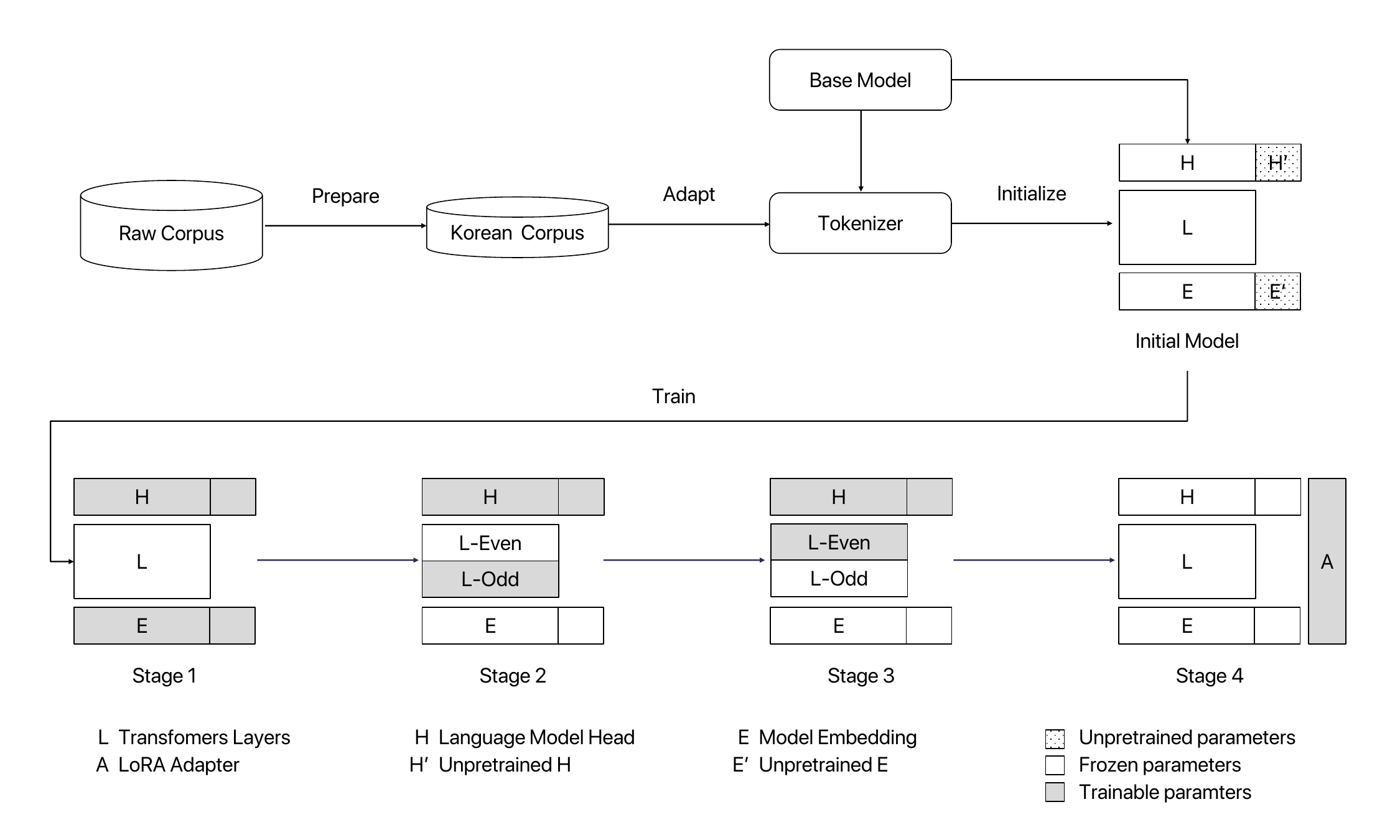}
\caption{Key steps in building RedWhale.}
\label{fig:fig_continual_pretraining_korean}
\end{figure}

Recent advancements in NLP have been significantly driven by the development and deployment of LLMs. These models have greatly enhanced our ability to understand and generate text, pushing the field closer to realizing Artificial General Intelligence (AGI) across multiple languages. However, most LLM development has concentrated on English, resulting in a marked deficiency in multilingual capabilities, particularly for non-English languages like Korean.

To address this gap and enhance Korean language processing, we introduce RedWhale\footnote{\url{https://huggingface.co/TwinDoc/RedWhale-tv-10.8B-v1.0}}, a pioneering model specifically designed for Korean. RedWhale integrates several significant improvements, including an extensive Korean corpus preprocessing pipeline, a specialized Korean tokenizer, effective model initialization, and a continuous multistage pretraining strategy. These innovations collectively enhance the model's ability to understand and generate Korean text, leading to higher efficiency and accuracy. Furthermore, they offer a cost-effective solution to improve NLP capabilities for the Korean language, making advanced language technology more accessible and effective.

Our research is inspired by and positioned within the context of preceding works such as Mistral 7B (\cite{jiang2023mistral}), which introduces grouped-query attention and sliding window attention for enhanced performance and efficiency over the well-known LLaMA2 model (\cite{touvron2023llama}). Additionally, the Efficient and Effective Text Encoding for Chinese LLaMA and Alpaca extends LLaMA's vocabulary to better accommodate Chinese texts (\cite{cui2024efficient}). The EEVE-Korean model showcases an effective vocabulary expansion method for Korean language processing (\cite{kim2024efficient}). Furthermore, the adaptation of LLaMA-2 into LLaMA-2-Ko, with an expanded vocabulary and Korean corpus\footnote{\url{https://huggingface.co/beomi/open-llama-2-ko-7b}}, and SOLAR 10.7B's introduction of depth up-scaling highlight the broader movement towards creating more inclusive and efficient LLMs that cater to a variety of languages beyond English (\cite{kim2023solar}).

The process of building a model typically involves two main phases: pretraining and model alignment, also known as post-training. Pretraining entails training the model on a large corpus of data to learn general patterns, structures, and representations. This phase equips the model with a broad understanding of language, enabling it to perform various tasks with a general baseline of competence. Model alignment, or post-training, follows pretraining, where the pretrained model is fine-tuned or adjusted to meet specific objectives, tasks, or ethical considerations.

In this study, an additional step was incorporated into the pretraining phase: language adaption. As illustrated in Figure~\ref{fig:fig_continual_pretraining_korean}, an English pretrained model (base model) was used as the starting point, and further pretraining was conducted to specifically adapt the model for Korean. This process resulted in a pretrained model optimized for Korean, which can serve as an optimal initialization for many Korean downstream tasks.

When addressing continuous pretraining, two main concerns typically arise: avoiding catastrophic forgetting and achieving knowledge transfer to enhance end-task performance. Given our focus on achieving knowledge transfer to enhance end-task performance specifically in Korean language processing, we evaluated end-task performance within this context. To achieve this, we applied two types of metrics: Causal Language Modeling (CLM) metrics to monitor pretraining effectiveness and the KoBEST(~\cite{kim2022kobestkoreanbalancedevaluation}) to assess the model's ability in advanced Korean linguistic tasks.

Our contributions in this study include a delineated four-step process for effectively adapting an English LLM to Korean:

\paragraph{Enhancing Korean corpus quality} A meticulous preprocessing pipeline was implemented to refine the quality of the Korean corpus. This involved the judicious selection and refinement of data to eliminate noisy and inconsistent data, thereby reducing computational demand and enhancing training efficacy by focusing on high-quality data.

\paragraph{Adapting an efficient Korean tokenizer} An advanced tokenizer was developed to efficiently process Korean text. By adjusting the tokenizer's vocabulary size and analyzing tokens, an optimal balance between input complexity and embedding complexity was achieved. This involved training a SentencePiece model on the Korean corpus and removing unnecessary tokens from the base tokenizer before combining them, resulting in a more efficient and effective tokenizer.

\paragraph{Initializing model weights effectively} Training expenses were minimized through optimal initialization methods, particularly for the Embedding and Language Model (LM) Head components. By maximizing the utilization of pretraining weights and carefully analyzing and calculating the weight of each newly created Korean token, computational efforts required in subsequent training steps were significantly decreased.

\paragraph{Implementing comprehensive multistage training} A four-stage training strategy was employed to accommodate hardware limitations and ensure effective pretraining. This strategy began with the initial training of the Embedding and LM Head components, followed by sequential adjustments to the pretrained Transformer blocks, also known as layers. Finally, Low-Rank Adaptation of Large Language Models (LoRA) (\cite{hu2021lora}) was applied across the model to achieve thorough alignment. After pretraining, model alignment was further refined using general instruction tuning and financial question-answering tuning to evaluate the model's quality for specific tasks.


\section{Related Works}
\label{sec:headings}

This section outlines the foundational background that informs the development of our model. We begin our review by examining foundational LLMs, including Llama2~\cite{touvron2023llama}, Mistral~\cite{jiang2023mistral}, and their derivatives. A focal point of our discussion is the Chinese variant of Llama2, proposed by Cui et al.~\cite{cui2024efficient}, which has undergone continuous pretraining from its original version, demonstrating improved comprehension and generation of Chinese text.

In addition, we explore open-llama-2-ko-7b\footnote{\url{https://huggingface.co/beomi/open-llama-2-ko-7b}}, an initial effort to tailor Llama2 for the Korean language, showcasing the model's adaptability and potential for language-specific enhancements. Further, we delve into the EEVE model (\cite{kim2024efficient}), which employs intricate and comprehensive techniques to adapt SOLAR for the Korean language (\cite{kim2023solar}). While these techniques are thorough, they may be considered overly complex for certain practical continual pretraining processes.

To provide a broader context, we also reference proprietary models by large companies such as GPT-4 (\cite{openai2024gpt4technicalreport}), Gemini (\cite{geminiteam2024geminifamilyhighlycapable}), Claude (\cite{anthropic2023gpt4technicalreport}), PalLM (\cite{anil2023palm2technicalreport}), and  HyperCLOVA X (\cite{yoo2024hyperclovaxtechnicalreport}). These models contain hundreds of billions of parameters and are pretrained from scratch on massive text data.

Furthermore, we acknowledge the fast-paced evolution of LLMs, with new findings, models, and techniques being published within months. Key surveys by Han et al.\cite{han2021pretrainedmodelspastpresent}, Liu et al.\cite{liu2021pretrainpromptpredictsystematic}, Zhao et al.\cite{zhao2023surveylargelanguagemodels}, Zhou et al.\cite{zhou2023comprehensivesurveypretrainedfoundation}, and Yıldız et al.~\cite{yıldız2024investigatingcontinualpretraininglarge} provide a comprehensive overview of these rapid advancements.

Substantial research specifically focuses on continual pretraining, including works by Yıldız et al.\cite{yıldız2024investigatingcontinualpretraininglarge}, Ke et al.\cite{ke2023continualpretraininglanguagemodels}, and Xie et al.~\cite{xie2023efficientcontinualpretrainingbuilding}.

For our training process, we applied FlashAttention from Tri Dao et al.\cite{dao2022flashattentionfastmemoryefficientexact}, which uses IO-aware algorithms and tiling to optimize memory usage and training speeds in GPUs. Additionally, we utilized Gina Sohn et al.'s work on implementing scaled dot-product attention (SDPA) (\cite{sohn2024implementingoptimizingscaleddotproduct}), which demonstrates achieving full throughput with constant intermediate memory, offering an innovative approach to attention mechanisms in non-processor architectures.

By examining these works, we establish the context and motivation for our approach, highlighting the need for efficient and effective methods in the development of Korean language models.

\subsection{Foundational LLMs}
\paragraph{Llama2}

The introduction of the LLaMA 2 model represents a notable advancement in the landscape of LLMs, featuring configurations that span from 7 billion to 70 billion parameters. This development marks a considerable evolution beyond its predecessors. Among the primary innovations attributed to LLaMA 2 are a 40\% enhancement of the pretraining data corpus, a two-fold increase in the model's context length, and the adoption of grouped-query attention mechanisms. The developers have made available several variants of the model, including those with parameter sizes of 7B, 13B, and 70B. Comparative analyses suggest that LLaMA 2-Chat surpasses existing open-source conversational models in metrics of helpfulness and safety, positioning it as a competitive alternative to proprietary counterparts such as ChatGPT\footnote{\url{https://openai.com/blog/chatgpt}}, Gemini\footnote{\url{https://gemini.google.com}}, and Claude\footnote{\url{https://claude.ai}}. The emergence of LLaMA 2 and LLaMA 2-Chat underscores significant progress in artificial intelligence, offering robust models for both scholarly investigation and practical application. These models highlight a commitment to enhancing accessibility, safety, and the facilitation of community-driven enhancements.

\paragraph{Mistral}
Building on the success of LLaMA 2, Mistral 7B, a 7-billion-parameter language model, sets a new standard in NLP by outperforming the best available models, including the 13B-parameter LLaMA 2 across all benchmarks and the 34B-parameter LLaMA 1 in reasoning, mathematics, and code generation tasks. Engineered for both superior performance and efficiency, Mistral 7B incorporates grouped-query attention (GQA) and sliding window attention (SWA) mechanisms to enhance inference speed and manage longer sequences with lower computational costs. These innovations enable Mistral 7B to effectively handle high throughput and real-time applications. Additionally, a specialized version, Mistral 7B – Instruct, demonstrates superior capabilities in following instructions compared to similar models. Released under the Apache 2.0 license, Mistral 7B is designed for easy deployment and fine-tuning, showcasing its versatility and high performance across various tasks, including a chat model variant that surpasses existing benchmarks. This breakthrough signifies a promising direction toward developing affordable, efficient, and high-performing language models for a wide range of real-world applications.

\subsection{Derivatives}
\paragraph{SOLAR 10.7B}

The SOLAR 10.7B was introduced by Kim et al.(~\cite{kim2023solar}) as a derivative of Llama 2 and Mistral, with 10.7 billion parameters, demonstrating improved performance across a variety of NLP tasks. This performance is attributed to a novel upscaling technique called depth upscaling (DUS). Unlike previous scaling strategies, such as the Mixture of Experts (MoE) (\cite{cai2024surveymixtureexperts}), DUS emphasizes depthwise scaling and continuous pretraining, diverging from traditional methods by offering a simpler and more efficient upscaling process that does not require complex modifications for training and inference. The paper further reports that SOLAR 10.7B's effectiveness is enhanced in its variant, SOLAR 10.7B-Instruct, which is fine-tuned for executing complex instructions, thereby outperforming similar models like Mistral-8x7B-Instruct\footnote{\url{https://huggingface.co/mistralai/Mixtral-8x7B-Instruct-v0.1}}. An innovation of this method is the model's initialization from the Mistral 7B model weights under the Llama 2 architecture. This approach leverages the strengths of Mistral's robust model weights and the widespread support for the Llama 2 model. However, transitioning to a different architecture may result in the loss of some of Mistral's advantages, including GQA and SWA.

\paragraph{Chiness Llama2}

This model showcases the efforts to enhance the capabilities of LLaMA2 in understanding and generating Chinese by continuing its pretraining (\cite{cui2024efficient}). By adding an additional 20,000 Chinese tokens to LLaMA's vocabulary and utilizing methods such as secondary pre-training with Chinese data and fine-tuning with Chinese instruction datasets, the authors significantly enhance the model's proficiency in Chinese. This improvement allows the model to better understand and generate Chinese text and follow instructions more effectively. The paper presents experimental results that highlight the augmented model's improved performance on Chinese language tasks, delivering competitive outcomes even compared to larger models. Resources, including pre-trained models and training scripts, are made available on GitHub\footnote{\url{https://github.com/ymcui/Chinese-LLaMA-Alpaca-2}} to promote open research and encourage the adaptation of LLaMA to other languages. This serves as a crucial reference for guiding the development of our Korean LLM.

\paragraph{Open-Llama-2-Ko}

The Open-Llama-2-Ko represents a pioneering endeavor in the field of continual pretraining, specifically tailored for the Korean language. It builds upon the Llama 2 by incorporating an enhanced vocabulary and integrating a Korean corpus to augment its pretraining capabilities. This initiative places particular emphasis on the 7 billion and 13 billion parameter versions. A critical differentiator of the Open-Llama-2-Ko from the preceding Llama-2-Ko series is its exclusive reliance on publicly available Korean corpora for its dataset. It draws from esteemed sources such as AI Hub\footnote{\url{https://www.aihub.or.kr}}, Modu Corpus\footnote{\url{https://kli.korean.go.kr}}, and Korean Wikipedia\footnote{\url{https://ko.wikipedia.org}}, thereby underscoring its ability to enhance linguistic representation and model accessibility in the realm of Korean language processing.

\paragraph{EEVE}
This adaptation by Kim et al.~\cite{kim2024efficient} enhances LLMs for better understanding of both English and Korean texts, addressing the performance gap and inefficiencies faced by non-English languages in existing English-centric LLMs like SOLAR-10.7B and Phi-2. Utilizing an innovative Efficient and Effective Vocabulary Expansion (EEVE) approach that includes parameter freezing and subword initialization, this report challenges the prevailing belief that embedding new vocabulary requires extensive training tokens. It demonstrates significant improvements in non-English language proficiency with just 2 billion tokens. By January 2024, EEVE-Korean-10.8B-v1.0 had emerged as the leading Korean pre-trained model on Hugging Face's Open Ko-LLM Leaderboard, surpassing other models in Korean language tasks while maintaining robust English capabilities. The EEVE models, built on top of the latest English-centric LLMs and further optimized for Korean, effectively close the performance gap in computational efficiency and language proficiency for non-English languages. This work not only sets new benchmarks but also contributes to the open-source community, making these advancements accessible for broader research and application across various languages. However, the complex techniques introduced for adapting SOLAR to the Korean language may be deemed unnecessary for certain practical continual pretraining processes.

Building on comprehensive insights from foundational studies, this research is firmly rooted in previously discussed advancements and methodologies. The primary objective is to develop an efficient and high-performing large language model specifically tailored for the Korean language, with a focus on minimizing training costs without compromising quality. Inspiration is directly drawn from the Chinese variant of Llama2, which demonstrated significant improvements in understanding and generating Chinese text through continued pretraining and vocabulary expansion. Similarly, the Open-Llama-2-Ko project serves as a crucial reference point, highlighting the potential of integrating publicly available Korean corpora to enhance linguistic representation and model accessibility in Korean language processing.

Additionally, innovative approaches exemplified by the EEVE model are considered, which successfully addressed performance gaps and inefficiencies in non-English language processing within English-centric LLMs. Methods such as parameter freezing and subword embedding initialization were utilized in the EEVE approach, demonstrating that significant improvements in language proficiency could be achieved with a relatively small number of training tokens. Synthesizing these diverse strategies, our research aims to create a Korean LLM that is not only cost-effective but also maintains high accuracy and comprehension.

More importantly, our proposed process is flexible and robust, making it applicable to any new base model that emerges in the future. This adaptability ensures that our approach can continually integrate and leverage the latest advancements in LLM technology, providing a scalable solution for efficient language model training across diverse linguistic contexts.

\section{Proposed Method}
Our research presents an innovative contribution to the development of a continual pretraining approach tailored for the Korean language by outlining a comprehensive four-step process, as illustrated in Figure~\ref{fig:fig_continual_pretraining_korean}. This process encompasses (1) enhancing the quality of the Korean corpus, (2) efficient tokenizer adaptation for Korean, (3) effective model weight initialization, and (4) efficient multi-stage model training. Each step in this meticulous pipeline has been specifically crafted to construct high-quality models in a cost-effective manner. The subsequent sections provide a detailed description of each stage, highlighting the methodologies and techniques employed to achieve superior performance in Korean language processing.

\subsection{Enhancing Korean Corpus Quality}

At this phase, the project concentrated on refining a corpus sourced from publicly accessible Korean texts. The main goal was to decrease the corpus's size while concurrently improving its quality, a crucial factor for both cost-efficient and performance-efficient language model training (\cite{engstrom2024dsdmmodelawaredatasetselection, xie2023doremioptimizingdatamixtures, penedo2024finewebdatasetsdecantingweb}). To accomplish this, a detailed preprocessing pipeline was implemented, which included steps such as random selection, rule-based filtering, deduplication, and filtering based on perplexity.

Initially, a merged corpus was created by gathering Korean text data from sources such as AI Hub\footnote{\url{https://www.aihub.or.kr}} and our crawled data from the web. This merged data was relatively large, presenting a challenge: how to filter it to a smaller size while still maintaining quality. Therefore, a random selection technique was employed to significantly reduce the corpus's size. This method allowed for more efficient use of computational resources by focusing on a smaller, yet representative, subset of the original dataset.

For deduplication, an n-gram-based filtering method was used to ensure the dataset's uniqueness. This method analyzed the frequency of contiguous sequences of n-grams within the text to identify and eliminate duplicate or nearly identical content. Removing such redundancies was crucial to reduce the risk of overfitting the model on repetitive data.

The final step in the preprocessing sequence was perplexity-based filtering. Perplexity, a measure of the predictive accuracy of a probability model, was used to evaluate and filter documents based on their coherence and quality. This standard was vital for the development of language models, ensuring that only documents of high quality and coherence were included in the training process. Although the application of perplexity-based filtering is high precision, low recall, which means we might miss a substantial number of clear documents, all chosen documents are of high quality.

\subsection{Efficient Tokenizer Adaptation for Korean}

\begin{figure}[ht]
\centering
\includegraphics[width=1.00\linewidth]{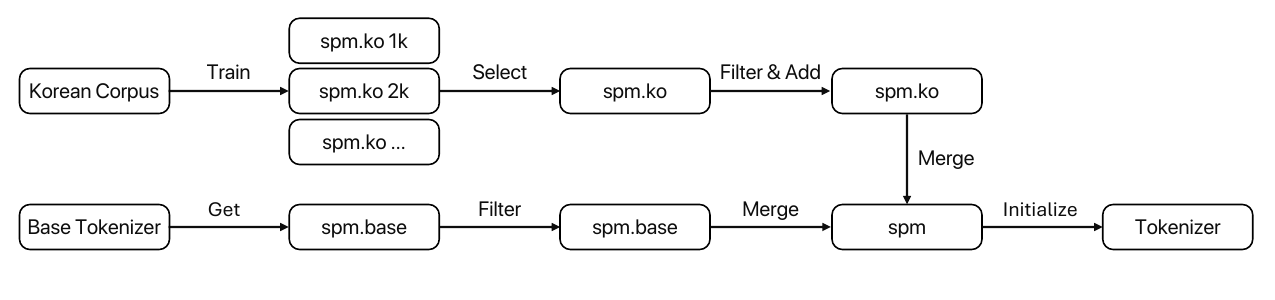}
\caption{Adapting the base tokenizer to Korean.}
\label{fig:tokenizer_adaption}
\end{figure}

The second step involves formulating a tokenizer specifically designed for the proficient processing of Korean textual data. The details of the tokenizer-building process are elucidated in Figure~\ref{fig:tokenizer_adaption}. The process entailed training a new SentencePiece model on a Korean corpus, applying rule-based token filtration, amalgamating it with the original tokenizer to form a hybrid tokenizer, and choosing an optimal vocabulary through iterative training and analysis.

Initially, Google SentencePiece\footnote{\url{https://github.com/google/sentencepiece}} was employed to train a SentencePiece model (SPM). Subsequently, rule-based token filtration was applied to both the newly trained SPM model and the base model, with a particular emphasis on processing Korean characters. The refined SPM model was then amalgamated with the original SPM model, resulting in a hybrid tokenizer. This hybrid tokenizer retained the ability to interpret English and other languages previously compatible with the pre-trained model while efficiently processing Korean text.

In training the new SPM model, the refined corpus obtained in the preliminary stages was utilized. This corpus was effectively reduced to an appropriate size without additional sampling, limiting maximum sentence length, or filtering character coverage. The vocabulary size was determined by balancing the trade-off between input complexity and model complexity. A larger vocabulary size generally reduces the number of tokens required by the model, thereby lowering input complexity, but it increases model complexity due to the expansion of embedding and LM Head dimensions.

Next, predefined Korean pieces were directly added to the trained SPM model. Following this, the refinement of both the base SPM model and the trained SPM model involved the removal of unnecessary characters from the trained SPM model to create a distilled, high-quality, and coherent tokenizer. The criteria for refinement were established through manual definitions, prioritizing tokens of the highest frequency.

Finally, the refined SPM models were merged to construct the final tokenizer.

This process involved multiple iterations of tokenizer training and analysis, ensuring that the newly formulated tokenizer encompassed a comprehensive and pertinent vocabulary for our project.

\subsection{Effective Model Weight Initialization}

\begin{figure}[ht]
\centering
\includegraphics[width=1.0\linewidth]{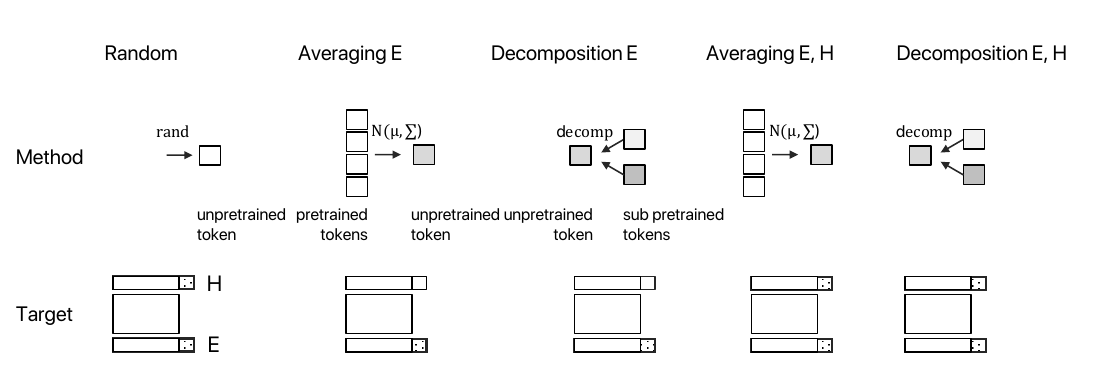}
\caption{Illustration of Five Initialization Methods.}
\label{fig:fig_init_methods}
\end{figure}

To initialize the model, the base model was adapted to accommodate the newly created Korean token embeddings produced by the tokenizer in the previous step. This adaptation required expanding both the model's Embedding (E) layer and the LM Head (H) to include the new Korean tokens while preserving the original model's architecture. The objective was to minimize training expenses through optimal initialization methods by maximizing the utilization of pretraining weights. Five initialization methods were experimented with, as illustrated in Figure~\ref{fig:fig_init_methods}.

\paragraph{Random Initialization:} 
HuggingFace's random initialization method\footnote{\url{https://github.com/huggingface/transformers.git}} generates new weights for tokens lacking pre-existing embeddings. This approach is particularly beneficial when no prior information or pretrained weights are available for specific tokens, providing a neutral starting point for the learning process during model training. While this method does not leverage any prior knowledge, it serves as a baseline for comparison with more sophisticated initialization techniques.

\paragraph{Averaging with Embedding:}
The Averaging with Embedding method utilizes the average of all pretrained weights from the entire vocabulary (pretrained tokens) to initialize the embeddings of new tokens (unpretrained tokens), thereby providing a more informed starting point that may lead to faster convergence during training. This approach involves calculating the average weights of the pretrained embeddings, identifying new tokens, and using these averaged weights to initialize the embeddings of the new tokens, based on the assumption that the embeddings of the new tokens are aligned with the distribution of the pretrained tokens. Specifically, during the initialization process, the method computes the mean and covariance of the original embeddings and samples from a scaled multivariate normal distribution to generate embeddings for the new tokens. By integrating pretrained knowledge with a probabilistic approach, this method improves the alignment between new and existing embeddings, thereby enhancing both the performance and efficiency of the model. An enhanced version of this method is presented in the study by Dobler et al.(\cite{Dobler_2023}), which transfers information from the source model's pretrained embedding matrix to a new embedding matrix for the target language. Instead of computing the distribution of the entire pretrained embedding and sampling for new tokens, the key idea is to use overlapping tokens between both tokenizers as anchor points and represent new target language tokens as a weighted combination of the overlapping tokens' embeddings.

\paragraph{Token Decomposition with Embedding:}
Token decomposition with embedding involves breaking down a new token into smaller sub-pretrained token components and averaging their embeddings to initialize the new token's embedding. This method assumes that an unpretrained token's embeddings can be captured by the averaged embeddings of its sub-pretrained tokens(~\cite{welch2020improvinglowcomputelanguage, kim2024efficient}). It is particularly useful for tokens that can be decomposed into meaningful subunits, such as morphemes or character n-grams. The implementation steps include decomposing each new token into its sub-pretrained token components, retrieving the embeddings for these sub-pretrained tokens from the pretrained embedding, computing the average of these subtoken embeddings, and initializing the new token's embedding with this average. This method leverages the compositional nature of language, where the meaning of a token can often be inferred from its parts.

\paragraph{Averaging with Embedding and LM Head:}
This approach extends the averaging method by including the weights of both the embeddings and the LM Head. By averaging the weights for both components, the alignment between the embedding layer and the head is improved, potentially enhancing the model's performance for new tokens. This dual averaging ensures consistency across different parts of the model.

\paragraph{Token Decomposition for Embedding and LM Head:}
This method extends the Token Decomposition with Embeddding approach by performing decomposition for both the embedding and the LM head. It ensures that new tokens are initialized in a manner that leverages both their subtoken structures and the pretrained knowledge for the LM head. This comprehensive approach aims to maximize the utilization of pretrained knowledge for improved model initialization.

Given that the input model in this study is in English and will be customized for Korean, we anticipate limitations in the effective splitting of tokens when employing Token Deposition methods. Consequently, this approach may not surpass the performance of averaging all embeddings or averaging similar tokens. To address this, we will systematically experiment with the five methods discussed, conducting comprehensive evaluations to identify the most suitable approach for this research. These diverse initialization strategies will be rigorously assessed to ensure the most effective integration of new Korean tokens, ultimately enhancing model performance and training efficiency.

\subsection{Effective Multi-Stage Model Training}


The training of LLMs presents two primary challenges: memory constraints, which frequently result in OOM errors, and the significant computational resources required for effective training. Recent advancements in the field have sought to address these challenges through several approaches. Firstly, efficient training methodologies have been developed to reduce memory consumption and accelerate computational processes. Notable implementations include FlashAttention2 (~\cite{dao2022flashattentionfastmemoryefficientexact}), SPDA\footnote{\url{https://pytorch.org/tutorials/intermediate/scaled_dot_product_attention_tutorial.html}}, and Unsloth\footnote{\url{https://unsloth.ai}}, which enhance backend performance. Secondly, techniques aimed at minimizing the memory footprint of model weights, such as QLoRA (\cite{dettmers2023qloraefficientfinetuningquantized}), LST (\cite{sung2022lstladdersidetuningparameter}), and QST (\cite{zhang2024quantizedtuningfastmemoryefficient}), as well as those that decrease the memory requirements of optimizer states, such as GaLore (\cite{zhao2024galorememoryefficientllmtraining}), and 8-bit-optimizer in BitsAndBytes\footnote{\url{https://github.com/bitsandbytes-foundation/bitsandbytes}}, have been introduced. Thirdly, parameter-efficient strategies, particularly LoRA, have been shown to significantly enhance memory and computational efficiency. LoRA achieves this by fine-tuning a limited subset of additional model parameters rather than the entire parameter set. However, it is important to note that this method is primarily suitable for fine-tuning due to the constraints of approximating full-rank matrices with low-rank representations. Finally, when computational resources are limited, partitioning the training process into smaller stages has proven to be an effective strategy (\cite{cui2024efficient}, \cite{kim2024efficient}).

Building on the insights provided by these foundational studies, a stepwise gradual adaptation approach is proposed to better align with hardware limitations and reduce overall training time. The proposed method consists of three stages: (1) Training new modules, specifically the Embedding and LM Head, using a large learning rate to facilitate rapid adaptation; (2) Tuning pre-trained modules, particularly the Transformer Blocks, with a smaller learning rate; and (3) Consolidating the entire model by applying LoRA, a parameter-efficient technique.

In the course of training, both FlashAttention2 and SPDA are utilized to minimize memory overhead and enhance processing speed, as these methods have demonstrated nearly equivalent performance.

The implementation of gradual adaptation through staged training, coupled with the application of training acceleration techniques, has proven effective in preserving overall model performance while successfully mitigating the risk of OOM errors.

Similar to Llama and Mistral-based models, a Transformers-based CLM model, was utilized. This model comprises three main components: Embedding, Transformer Blocks, and the LM Head. During the initialization stage, the Embedding and LM Head were adapted to fit the customized Korean tokenizer, while the Transformer Blocks were transferred intact from a pretrained model.

The training strategy was divided into three stages to ensure effective integration of new components, manage hardware limitations, and optimize computational resources. Initially, the focus was on the Embedding and LM Head components, which contained untrained weights. These components were trained first to rapidly integrate the new model parts. Next, the pretrained Transformer Blocks, which primarily contain the model's weights, were adjusted. Finally, the entire model was trained to ensure thorough alignment.

\paragraph{Embedding Adaptation} In the first stage, the emphasis was on the initial training of the Embedding and LM Head components. The embedding layer, responsible for converting input text into numerical representations, required the initialization of newly introduced Korean tokens. This training ensured that these new embeddings accurately represented the Korean text in the model's latent space. Concurrently, the LM Head, which generates the model's output, was trained to align its new weights with the pretrained components, thereby ensuring coherent output generation for Korean text. This rapid integration of new components allowed the model to start processing and generating Korean text accurately from the early stages, facilitating a more controlled and efficient adaptation of the new tokens into the existing model architecture.

\paragraph{Layer Adaptation} The second stage involved adjusting the pretrained Transformer Blocks to accommodate the changes introduced by the new embeddings and LM Head. Transformer Blocks, forming the core of the model, process input text through attention mechanisms and other complex transformations. Rather than fine-tuning these blocks all at once, a staged approach was employed, possibly by freezing certain layers and training others sequentially to efficiently manage computational resources. Specifically, the training of the Transformer layers was divided into two sub-stages. Initially, the layers in odd positions (1, 3, 5) were trained, followed by the training of layers in even positions (0, 2, 4). This approach offers two main benefits: it can be performed with limited GPU memory, and it allows each part of the model to learn distinct portions of the data.

\paragraph{Model Consolidation} The final stage involved comprehensive training of the entire model to ensure thorough alignment and optimal performance. Once the new components were integrated and the Transformer Blocks adjusted, the entire model was trained holistically. This ensured that all parts of the model, including Embeddings, Transformer blocks (Layers), and the LM Head, were aligned and worked cohesively. Tuning across all layers was conducted to harmonize the interactions between different model components. Techniques such as LoRA were employed to make final adjustments and enhance the model's efficiency.

\subsection{Evaluation}

In this research, model evaluation is conducted continuously throughout the entire model-building process, encompassing initialization, staged pretraining, and task-specific fine-tuning.

\paragraph{Pretraining Evaluation} The evaluation during pretraining focuses on key metrics derived from the CLM learning objective, including training loss, evaluation loss, and evaluation accuracy for the next-token prediction task. These metrics collectively reflect the effectiveness of the initialization and pretraining phases, enabling the detection of potential abnormalities early in the training process. By closely monitoring these indicators, we can promptly address any issues, ensuring that the model learns effectively from the outset.

\paragraph{Evaluation on General Korean Tasks} The model's performance on Korean language tasks is evaluated through a fine-tuning step using the kollm-conversations dataset\footnote{\url{https://huggingface.co/datasets/davidkim205/kollm-converations}} following the initial pretraining phase. Subsequently, four tasks from the KoBEST benchmark—BoolQ, COPA, HellaSwag, and SentiNeg—are employed to assess the model's quality and proficiency in handling advanced Korean linguistic challenges (\cite{kim2022kobestkoreanbalancedevaluation}).

The kollm-conversations dataset is a publicly available, general-purpose dataset specifically designed for Supervised Fine-Tuning (SFT). It comprises an integrated collection of conversations leveraging existing Korean datasets from platforms like Hugging Face and GitHub.

The four tasks in KoBEST, developed by professional Korean linguists, offer a robust framework for evaluating and comparing the quality of our model against reference models. This approach allows us to determine whether our model's performance is directly comparable to previously published results, thereby providing valuable insights into its effectiveness in Korean language processing.

\paragraph{Financial QA Evaluation} To assess the model's capabilities within the financial domain, we utilize proprietary Financial QA datasets specifically curated for commercial applications in the finance sector. These datasets, generated with the assistance of ChatGPT and meticulously post-processed for high quality, consist of context excerpts from longer documents paired with corresponding questions and answers. The questions are designed to challenge the model's abilities in tasks such as summarization, recognition, and data extraction. The Financial QA datasets are divided into two collections: QA-1, which includes 5,210 training samples and 50 test samples, and QA-2, which contains 6,123 training samples and 70 test samples. Both collections are tailored for research and analysis within the financial sector. The model's performance is evaluated using the Accuracy Confidence Metric (ACM), a method that rigorously analyzes the predicted answers and categorizes them into three groups: correct (indicating an exact match or very high confidence), partially correct (indicating a partial match or moderate confidence), and incorrect. The ACM provides a comprehensive evaluation by considering both the accuracy of the answers and the model's confidence level, offering a nuanced understanding of its capabilities in financial contexts.

In conclusion, the evaluation methodology adopted in this research provides a comprehensive and continuous assessment of the model's performance across different stages of its development. During the pretraining phase, key metrics such as training loss, evaluation loss, and evaluation accuracy are closely monitored to ensure the model's initial learning process is effective and any issues are identified early. The model's proficiency in handling general Korean language tasks is rigorously tested through fine-tuning on the kollm-conversations dataset and benchmarking against tasks from the KoBEST framework. This allows for a clear comparison with established models, ensuring our model meets or exceeds the expected standards in Korean linguistic challenges. Finally, the specialized Financial QA evaluation demonstrates the model's capability in applying its knowledge within the financial domain, with a detailed analysis provided by the ACM. This multi-faceted evaluation strategy ensures that the model is not only well-prepared for general language tasks but also excels in domain-specific applications, particularly in the finance sector.

\section{Experimental Results}\label{sec:experimental_results}

\subsection{Corpus}



\begin{CJK}{UTF8}{mj}
\begin{table}[t]
\centering
\begin{minipage}[t]{0.45\textwidth}
\centering
\caption{Raw and processed corpus}
\label{table:aihub-corpus}
\begin{tabular}{llr}
\toprule
 & Source & Size (GB) \\
\midrule
AI HUB data & AI HUB & 43.9 \\
Web-crawled data & Mixed sources & 77.7 \\
\midrule
Raw corpus& AI HUB & 121.6 \\
\midrule
Processed corpus (dataset)& AI HUB & 43.9 \\
\bottomrule
\end{tabular}
\end{minipage}
\hfill
\begin{minipage}[t]{0.45\textwidth}
\vspace{0.7mm}
\centering
\caption{Train and eval dataset}
\label{table:train-test-dataset}
\begin{tabular}{lrr}
\toprule
           & Train     & Test \\
\midrule
\# Docs    & 34,586,285 & 1,000 \\
\# Tokens  & 9,634,172,928 & 4,280,320  \\
\# Samples & 2,352,093   & 1,045 \\
\bottomrule
\end{tabular}
\end{minipage}

\end{table}
\end{CJK}

The preprocessing pipeline for the Korean corpus involved several steps to ensure high-quality and coherent text, thereby enhancing training efficiency and model effectiveness. These steps included random selection, rule-based filtering, n-gram-based deduplication, and perplexity-based filtering. Table ~\ref{table:aihub-corpus} provides an overview of the Korean corpus data obtained from AI HUB and our web-crawled data, which serve as the foundational datasets for this study. Initially, the raw data corpus totaled 121.6GB, but after preprocessing, it was refined to 43.9GB, retaining only high-quality text for effective language model training.

The refined corpus was then split into training and evaluation sets, as shown in Table ~\ref{table:train-test-dataset}. The training set comprises approximately 34.6 million documents, containing a total of about 9.7 billion tokens, as counted by our customized Korean tokenizer. With an input context length of 4,096 tokens, this resulted in 2.4 million samples. This extensive dataset ensures the model is exposed to a wide variety of contexts, which is crucial for robust language model development.

The test set, although smaller, is designed to effectively evaluate the model's performance on the CLM next token prediction task in terms of loss and accuracy. It consists of 1,000 documents, including 4.3 million tokens, also counted using our customized Korean tokenizer, resulting in 1,045 samples. This random test set enables a precise assessment of the model's loss, accuracy, and generalization capabilities throughout the training process.

Although a formal evaluation using quantitative measures to assess the processed data's specific impact on model training compared to the original data has not yet been conducted, it is hypothesized that selecting and refining data to enhance each data sample's consistency and quality may lead to improved computational efficiency. Specifically, reducing data size by removing noisy and non-uniform samples can decrease computing costs by reducing resource requirements. This approach may also improve the training quality of the model by focusing on high-quality data, thereby optimizing the learning process.

\subsection{Tokenizer}




\begin{figure}[ht]
\centering
\begin{minipage}[t]{0.45\textwidth}
\centering
\includegraphics[width=\linewidth]{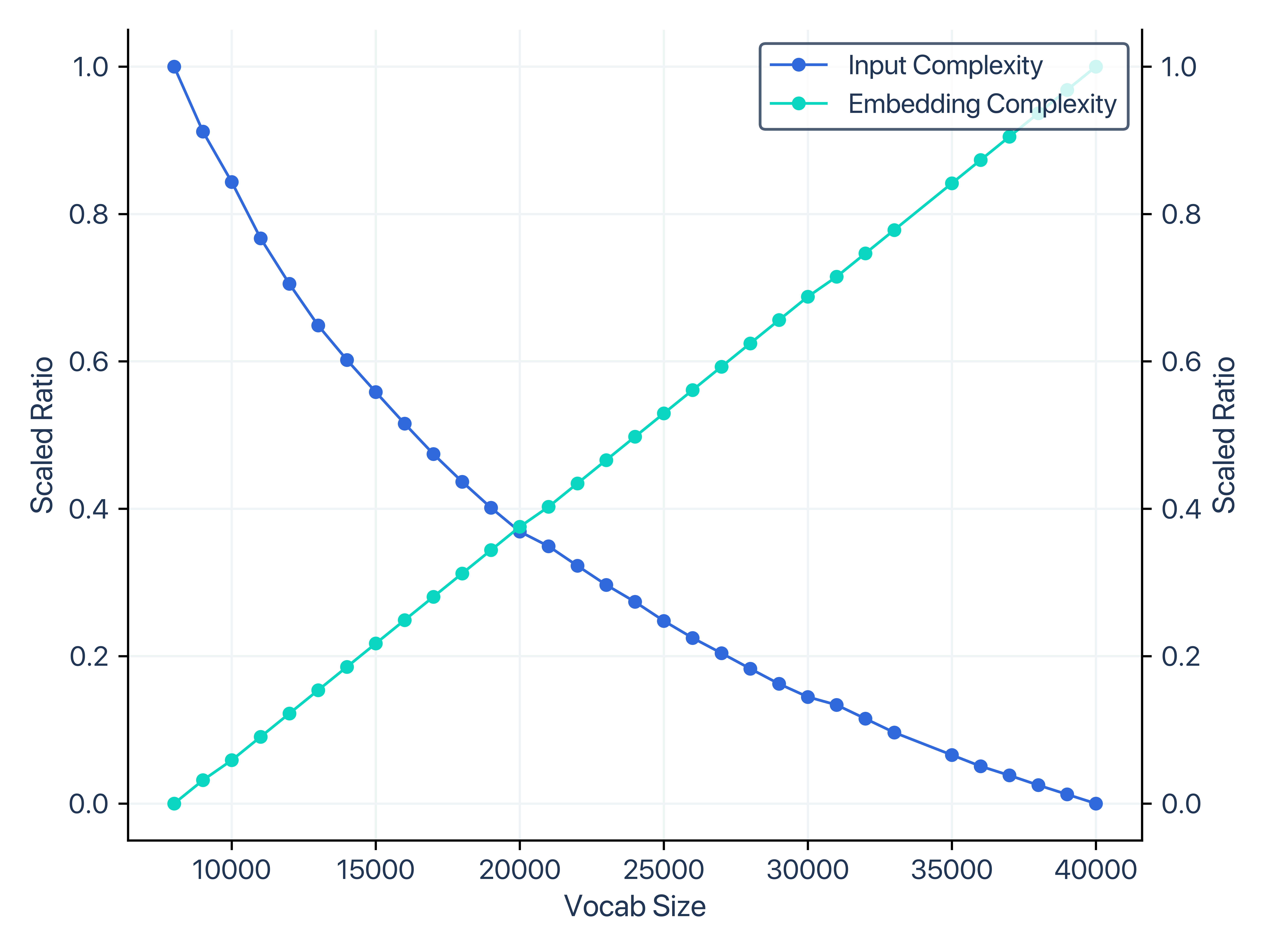}
\caption{Model complexity analysis.}
\label{fig:model_complexity_analysis}
\end{minipage}
\hfill
\begin{minipage}[t]{0.45\textwidth}
\centering
\includegraphics[width=\linewidth]{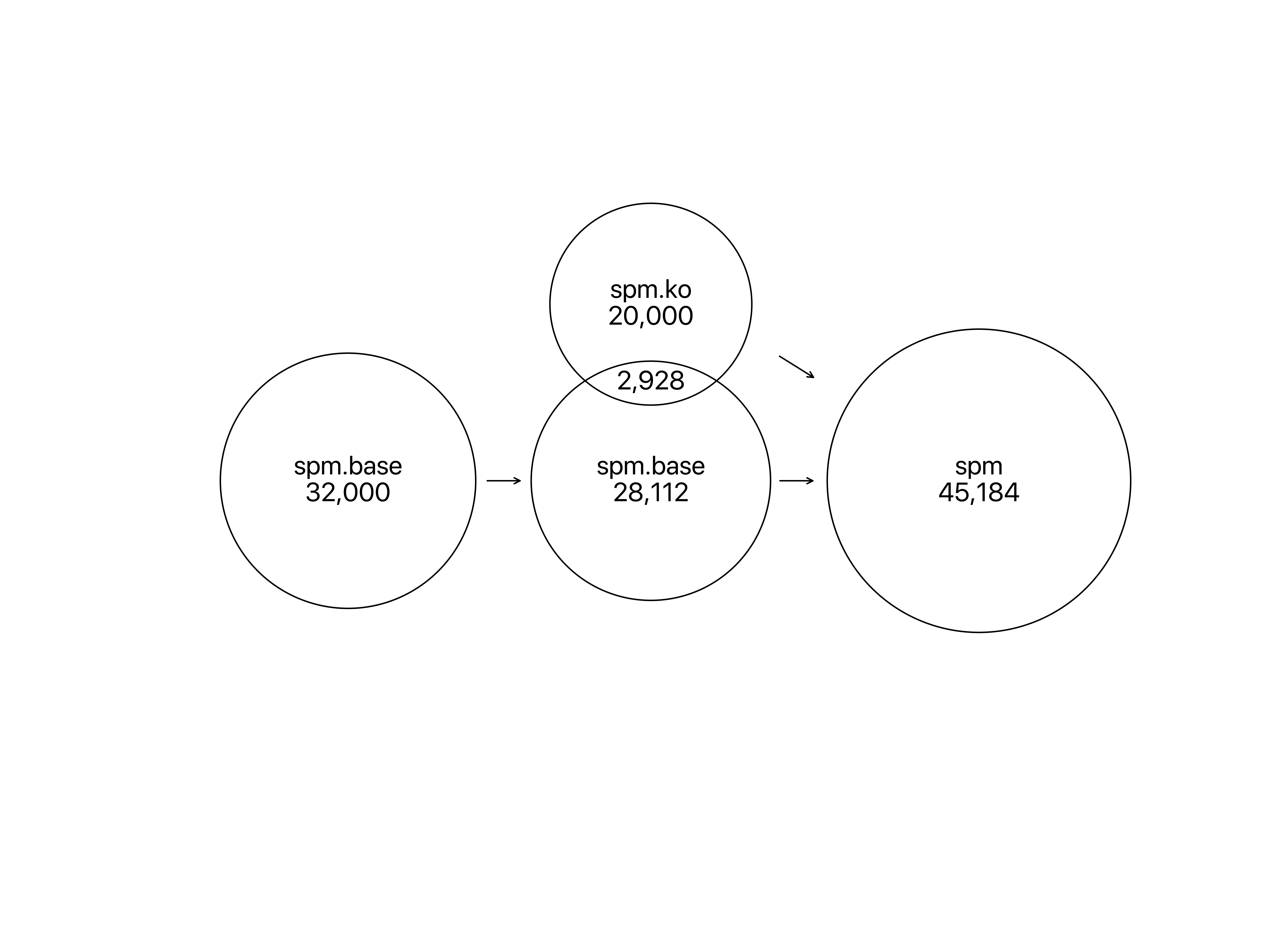}
\caption{Tokenizer adaption.}
\label{fig:tokenizer_creation}
\end{minipage}
\end{figure}

\begin{figure}[ht]
\centering
\includegraphics[width=0.8\linewidth]{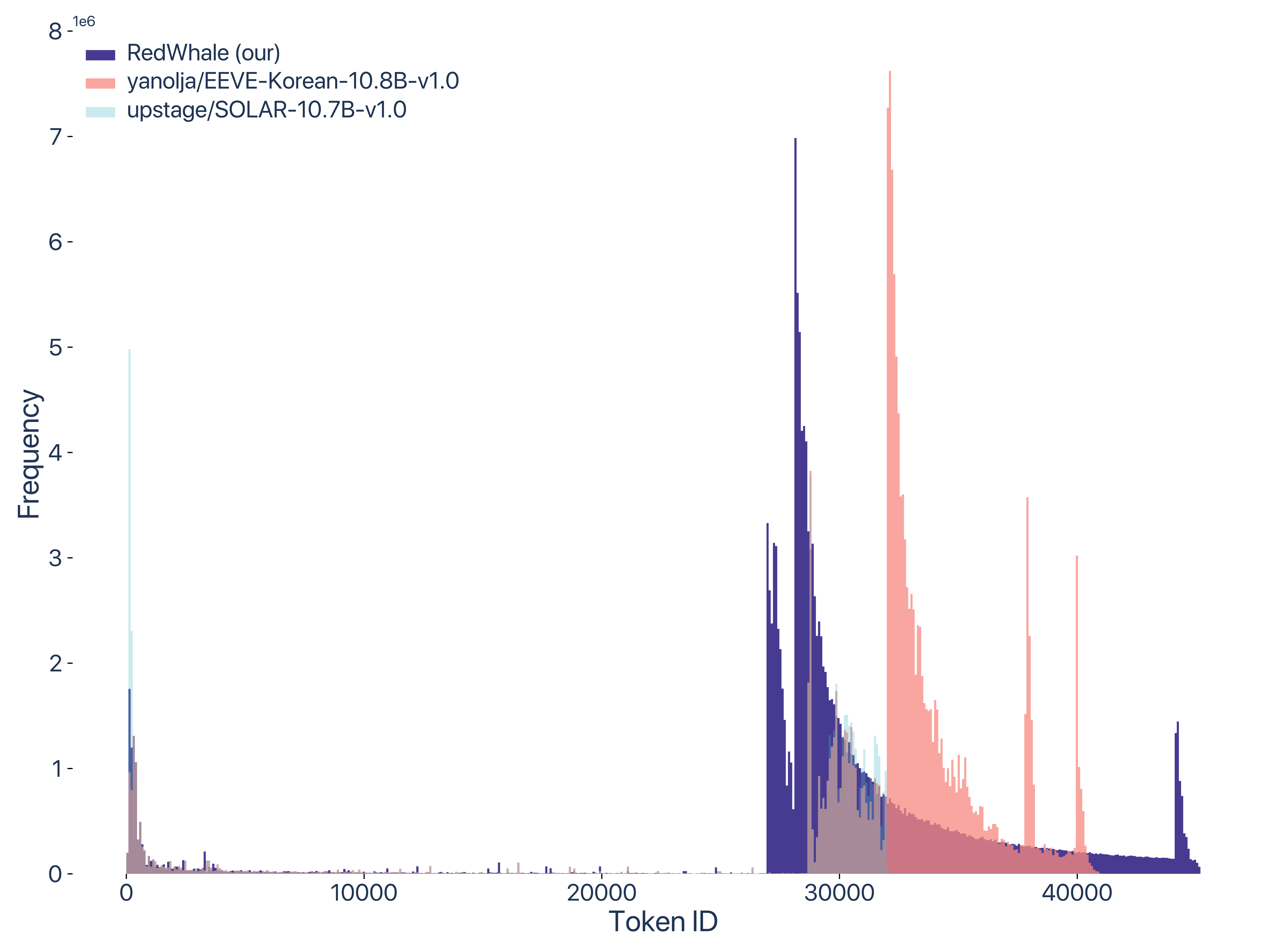}
\caption{Histogram of Token Frequency Distribution}
\label{fig:fig_tokenizer_histogram_of_token_frequency_distribution}
\end{figure}


As illustrated in Figure~\ref{fig:fig_continual_pretraining_korean}, the SOLAR model (\cite{kim2023solar}), an English-centric model, is used as the base model. To create an effective tokenizer for Korean data, we trained a SentencePiece model, adjusted the vocabulary by adding and removing necessary tokens, and then combined this new tokenizer with the base one. Our main objective was to determine the optimal vocabulary size, with experiments conducted on sizes ranging from 8,000 to 40,000 tokens.

Two key measures were used to evaluate the tokenizer: the Ratio of Input Complexity and the Ratio of Embedding Complexity. The Ratio of Input Complexity is calculated by dividing the number of tokens produced by the new tokenizer by the number of tokens produced by the base tokenizer. A lower ratio indicates reduced input complexity, allowing the model to process data faster and handle longer strings. The Ratio of Embedding Complexity is calculated by dividing the number of parameters in the new embedding model by the number of parameters in the base model's embedding. A lower ratio suggests a more efficient model with less memory and computation consumption.

The goal was to find a balance between these measures, reducing the Ratio of Input Complexity without significantly increasing the Ratio of Embedding Complexity. The experimental results are visualized in Figure~\ref{fig:model_complexity_analysis}. The experiments revealed that a vocabulary size of 20,000 was optimal for the Korean corpus used.

Figure~\ref{fig:tokenizer_creation} illustrates the final tokenizer, which is constructed by integrating the base tokenizer with the newly trained tokenizer. Notably, the overlap between tokens in the base tokenizer (trained on English) and the new tokenizer (trained on Korean) is minimal, consisting of only 2,928 tokens. This limited intersection highlights the enhanced capacity of the new tokenizer to process Korean data; however, it also underscores the substantial training required for the newly introduced tokens. The strategies for effective weight initialization for these untrained tokens will be explored in Subsection~\ref{subsec:effective_model_weight_initialization}.

Figure~\ref{fig:fig_tokenizer_histogram_of_token_frequency_distribution} shows histograms of token frequency distribution across three models, tested with one million Korean documents. The x-axis represents the Token ID, and the y-axis represents the frequency. The SOLAR Model (English-centric base model) displays a scattered distribution of token IDs from 0 to 32K, with a high concentration of tokens with very low Token IDs. Some spikes in frequency at higher Token IDs around 30K suggest frequent use of certain character-based and rare Korean-specific tokens. The EEVE tokenizer and our tokenizer show a distinct token usage pattern with pronounced peaks, indicating effective adaptation for Korean data. Most tokens are concentrated in the 30K to 40K range for EEVE and the 30K to 45K range for our model, where new Korean tokens have been added.

The analysis demonstrates that adding Korean tokens to the model is more effective than using an English-centric model. However, the concentration of tokens in the newly added ranges requires additional training to adapt to these new tokens, as there is less knowledge to transfer from the base model.

\subsection{Initialization}
\label{subsec:effective_model_weight_initialization}


\begin{table}[htbp]
\centering
\caption{Detailed comparison of initialization methods}
\label{table:comparison-initialization-method}
\begin{tabularx}{\textwidth}{p{2.9cm}<{\raggedright} p{2.7cm}<{\raggedright} *{8}{>{\raggedleft\arraybackslash}X}}
\toprule

\multicolumn{3}{l}{\textbf{Method}} & \multicolumn{2}{l}{\textbf{CLM Metrics}} & \multicolumn{5}{l}{\textbf{KoBEST}} \\
\cmidrule(r){1-3} \cmidrule(r){4-5} \cmidrule(r){6-10}

\textbf{Name} & \textbf{Source} & \textbf{Target} & \textbf{Loss} & \textbf{Accuracy} & \textbf{BQ} & \textbf{CP} & \textbf{HS} & \textbf{SN} & \textbf{AVG}\\
\midrule
Random & random & E, H & 17.5061 & 0.0036 & 0.5021 & 0.4730 & 0.2380 & 0.5063 & 0.4299\\
Averaging E & pretrained & E & 18.8953 & 0.0039 & 0.5021 & 0.4750 & 0.2460 & 0.4987 & 0.4305\\
Decomposition E & pretrained\_subtokens & E & 18.2258 & 0.0043 & 0.5021 & 0.4690 & 0.2400 & 0.5063 & 0.4294\\
Averaging E, H & pretrained & E, H & \textbf{10.6275} & 0.0736 & 0.5021 & 0.4700 & 0.2540 & 0.4937 & 0.4300\\
Decomposition E, H & pretrained\_subtokens & E, H & 11.0905 & \textbf{0.0930} & \textbf{0.5021} & \textbf{0.4860} & \textbf{0.2600} & \textbf{0.5264} & \textbf{0.4436}\\
\bottomrule
\end{tabularx}
\end{table}

The experimental results on initialization methods, presented in Table \ref{table:comparison-initialization-method}, offer a comprehensive comparison of various approaches used for token embeddings. The methods evaluated include Random Initialization, Averaging with Embedding (E), Token Decomposition with Embedding (E), Averaging with Embedding and LM Head (E, H), and Token Decomposition for Embedding and LM Head (E, H). Each method's performance is assessed across multiple metrics: Loss, Accuracy, and task-specific indicators such as BoolQ (BQ), COPA (CP), HellaSwag (HS), and SentiNeg (SN).

Random Initialization serves as the baseline. Despite not having the highest loss (17.5061) on the evaluation dataset, its performance in next-token prediction accuracy (0.0036) and individual tasks (BQ: 0.5021, CP: 0.4730, HS: 0.2380, SN: 0.5063) demonstrates limited effectiveness due to the lack of pretrained knowledge.

Averaging with Embedding (E) leverages the average of all pretrained weights. This method does not show an improvement in loss (18.8953) over random initialization but offers a slight improvement in accuracy (0.0039) and individual tasks (CP: 0.4750, HS: 0.2460).

Token Decomposition with Embedding (E), which decomposes new tokens into sub-pretrained components, achieves slightly better performance than the averaging method, with a loss of 18.2258 and accuracy of 0.0043. This method's performance on CP (0.4690) and SN (0.5063) suggests some potential benefits of leveraging subtoken structures, yet it remains close to the baseline in overall effectiveness.

Averaging with Embedding and LM Head (E, H) shows a significant reduction in loss (10.6275) and a substantial increase in accuracy (0.0736). This indicates that considering the LM head weights alongside the embedding weights leads to better alignment within the model, resulting in improved initialization. However, task-specific performance on BQ (0.5021), CP (0.4700), and HS (0.2540) remains only modestly improved.

Token Decomposition for Embedding and LM Head (E, H) emerges as the most effective method, with the lowest loss (11.0905) and the highest accuracy (0.0930). It also achieves the best performance across all individual tasks: BQ (0.5021), CP (0.4860), HS (0.2600), and SN (0.5264). This method’s comprehensive approach of leveraging both subtoken structures and pretrained LM head knowledge maximizes the utilization of existing knowledge, leading to superior model initialization and performance.

In summary, the experimental results clearly indicate that Token Decomposition for Embedding and LM Head (E, H) is the most effective method for initializing new token embeddings. This method outperforms others in terms of both general metrics (loss and accuracy) and task-specific performance. The strength of this approach lies in its ability to incorporate detailed subtoken information and pretrained LM head weights, providing an optimal starting point for model training.

\subsection{Training}
\label{subsec:training}

\paragraph{Hyperparamters}
\begin{table}
\centering
\caption{Hyperparamter details for experiments}
\label{table:train_hypeparams}
\begin{tabular}{llllllll}
\toprule
           &  & \multicolumn{2}{l}{Embedding, Head} & \multicolumn{2}{l}{Layers} &  & Entire \\
\cmidrule(r){3-4} \cmidrule(r){5-6} \cmidrule(r){7-8}
           &  & E', H' & E, H, E', H' & Odd & Even & & LoRA \\
\midrule
Exp 1 & Train tokens &  & 3.9B & 2.0B & 1.2B &  & 2.6B\\
        & Context length  &  & 4k   & 4k   & 4k   &  & 4k\\
        & Learning rate   &  & 100e-6 & 15e-6 & 15e-6 & & 100e-06\\
        & Effective batch  &  & 16  & 16  & 16  &  & 16\\
        & Trainable params(\%)  &  & 3.41  & 50.00  & 50.00  &  & 4.60\\
        & Train epoch     &  & 1  & 1  & 1  &  & 1\\
        & Actual train steps     &  & 59040  & 30240  & 18720  &  & 40320\\
\midrule
Exp 2 & Train tokens & 0.8B & 3.1B & - & - &  & -\\
        & Context length  & 4k   & 4k   & -   & -   &   & -\\
        & Learning rate   & 100e-6 & 100e-6 & - & - &   & -\\
        & Effective batch  & 16 & 16  & -  & -  &  & -\\
        & Trainable params(\%)  &  1.29 & 3.41  & -  & -  &  & -\\
        & Train epoch     & 1 & 1 & - & - &   & -\\
        & Actual train steps     & 12960 & 46080 & - & - &   & -\\
\bottomrule
\end{tabular}
\end{table}


\begin{figure}[ht]
\centering
\begin{minipage}{0.45\textwidth}
\centering
\includegraphics[width=\linewidth]{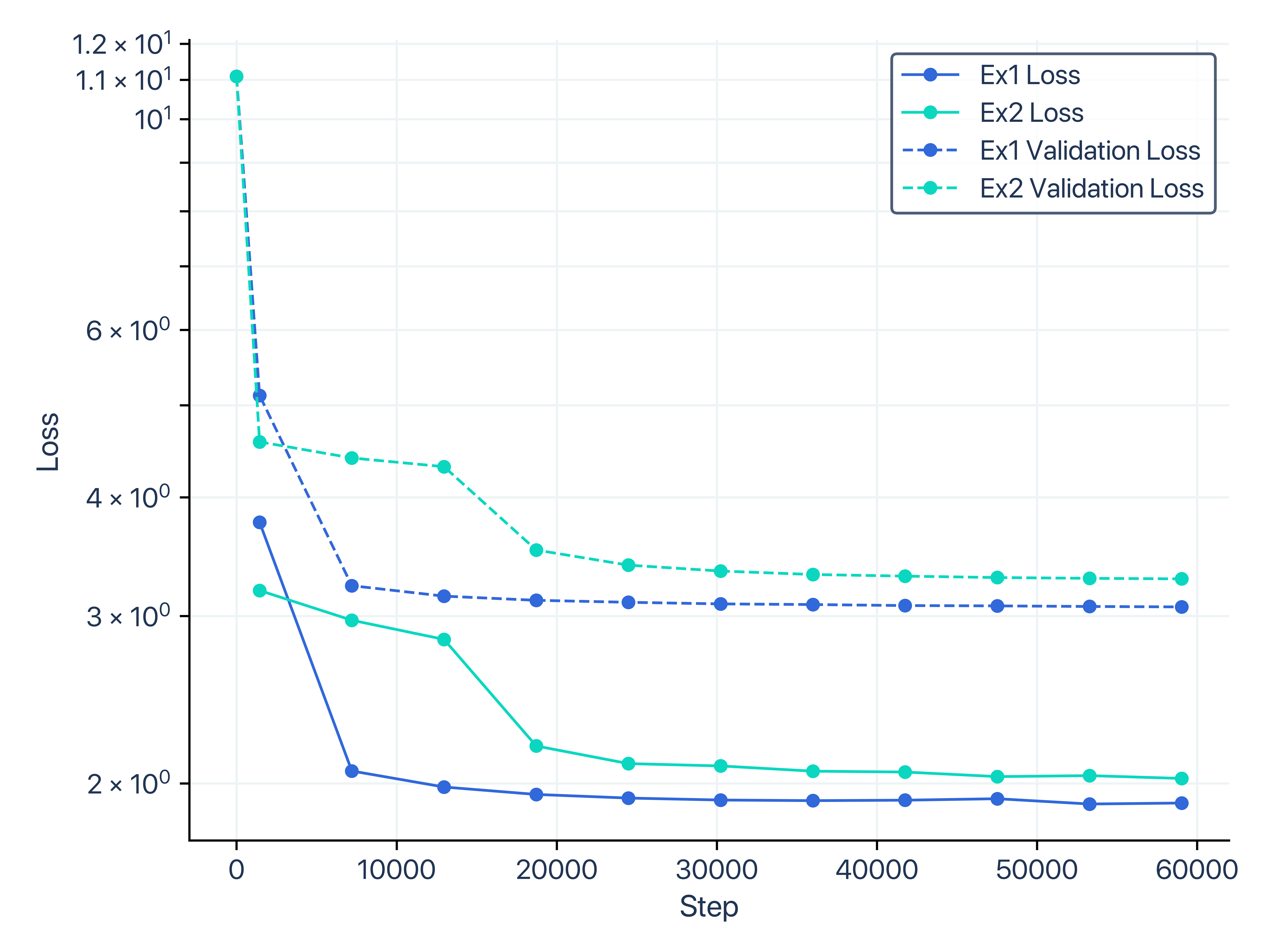}
\caption{Loss comparison of Ex1 and Ex2 during the Embedding and LM Head training stage.}
\label{fig:train_comparision_pt_metrics_loss_trials}
\end{minipage}%
\hspace{0.01\textwidth} 
\begin{minipage}{0.45\textwidth}
\centering
\includegraphics[width=\linewidth]{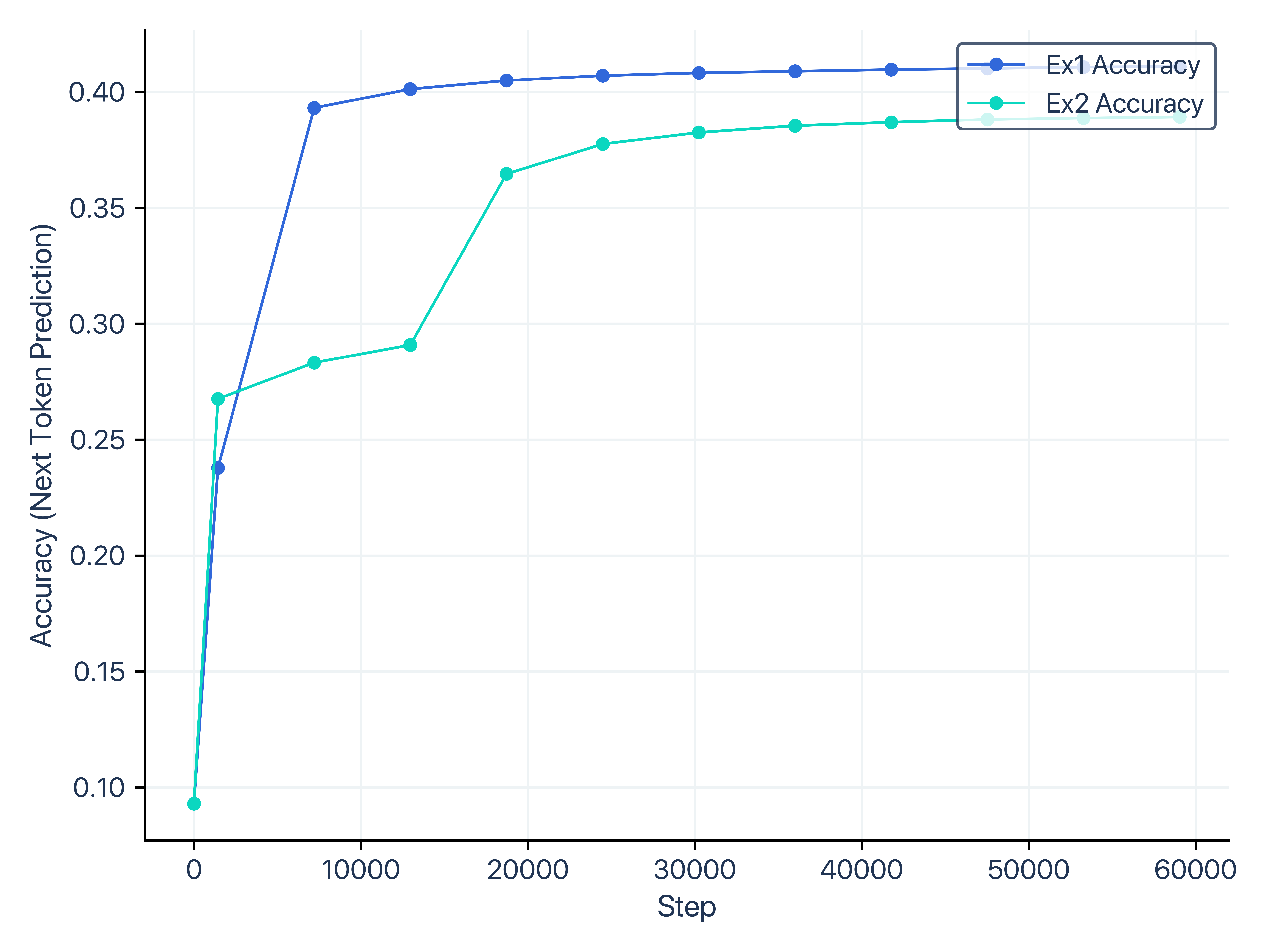}
\caption{Accuracy comparison of Ex1 and Ex2 during the Embedding and LM Head training stage.}
\label{fig:train_comparision_pt_metrics_accuracy_trials}
\end{minipage}
\end{figure}

\begin{figure}[ht]
\centering
\begin{minipage}{0.45\textwidth}
\centering
\includegraphics[width=\linewidth]{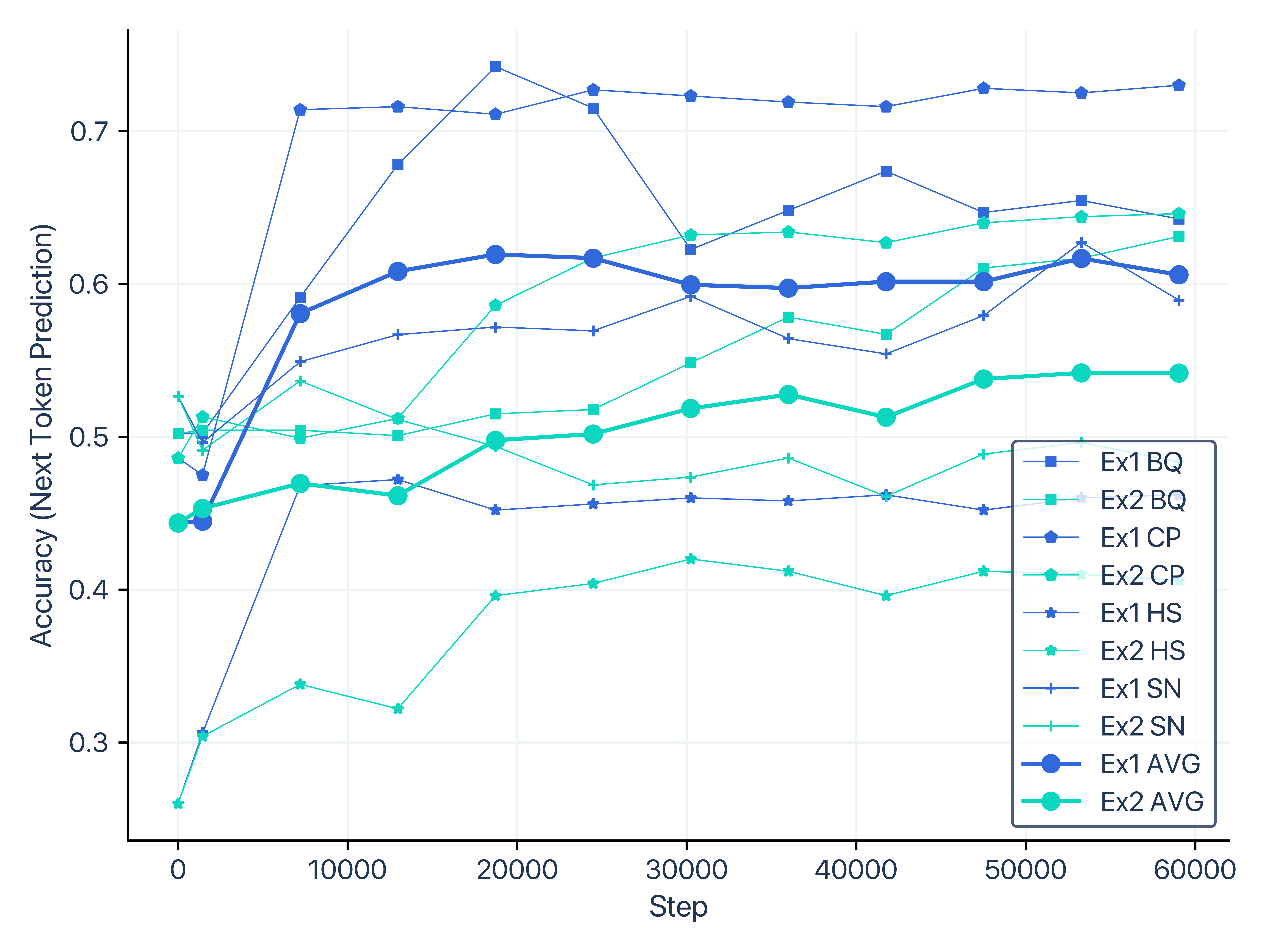}
\caption{Comparison of KoBEST between Ex1 and Ex2 during the Embedding and LM Head training stage.}
\label{fig:train_comparision_pt_metrics_kobest_trials}
\end{minipage}%
\hspace{0.01\textwidth} 
\begin{minipage}{0.45\textwidth}
\centering
\includegraphics[width=\linewidth]{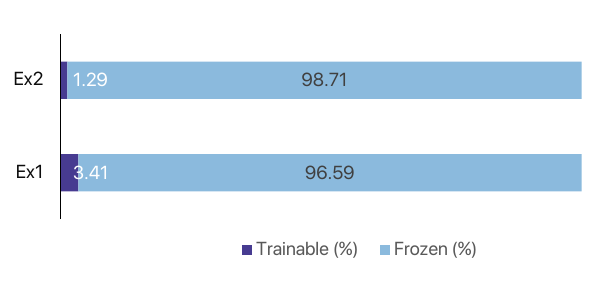}
\caption{Comparison of trainable parameters between Ex1 and Ex2 in Embedding and LM Head training.}
\label{fig:train_comparision_pt_params_trials}
\end{minipage}
\end{figure}

The model was trained using the default Torch AdamW optimizer, employing a cosine learning rate schedule with a warmup ratio of 0.03. Each stage was set to run for one epoch, although metrics were continuously monitored, and early stopping was implemented if performance converged.

Table ~\ref{table:train_hypeparams} provides detailed information on the hyperparameters used in our experiments. The primary parameter of interest was the learning rate, which was adjusted based on specific considerations. Firstly, the Embedding and LM Head required a relatively high learning rate (100µ) due to the presence of unpretrained weights. Secondly, the Transformer Blocks, which were transferred intact from the base pretrained model, necessitated a smaller learning rate (15µ). Lastly, when training the entire model using LoRA, a larger learning rate (100µ) was utilized to ensure optimal adaptation and performance of the modules initialized with LoRA. This careful adjustment of learning rates was crucial for maximizing the model's performance across different components and training stages.

\paragraph{Ablation Study}

The initial phase in the training pipeline involves training the Embedding and LM Head. The embedding includes components $E$ and $E'$, while the LM Head encompasses components $H$ and $H'$. In this context, $E$ and $H$ represent pretrained weights transferred from the base model, whereas $E'$ and $H'$ are initialized using a token decomposition method designed to leverage the pretrained weights effectively.

By following this three-main-stage training strategy, where each stage can be divided into sub-stages, to determine the effectiveness of the advanced Embedding and LM Head training strategy, two experiments (Ex) were conducted:

\begin{itemize}
\item Ex1: Entire Embedding and LM Head training. Training the entire set of weights $(E, E', H, H')$ from the start.
\item Ex2: Incremental training of Embedding and LM Head. Initially training the newly initialized weights $(E', H')$, followed by training the entire set $(E, E', H, H')$. In this trial, the training of the Embedding and Language Model (LM) Head was divided into two phases. First, only the untrained tokens were trained. After this initial step, the entire Embedding and LM Head were trained together. This approach aimed to test the effectiveness of quickly adapting the initialized model with a minimal number of training steps before proceeding to train the entire embedding and LM. This method was also mentioned in the EEVE study~\cite{kim2024efficient}.
\end{itemize}

\textit{Model performance}

Figure~\ref{fig:train_comparision_pt_metrics_loss_trials} illustrates the experimental results. Overall, Ex1 demonstrates superior training and evaluation outcomes in terms of both loss reduction and accuracy improvement compared to Ex2. Specifically, Ex1 exhibits lower training and evaluation loss, indicating enhanced model performance and convergence. Additionally, the evaluation accuracy, as shown in Figure~\ref{fig:train_comparision_pt_metrics_accuracy_trials} of Ex1, stabilizes at a higher value, suggesting better generalization.

During the very early training steps (0 to 1440), the loss reduction in Ex2 is more pronounced than in Ex1. This occurs because Ex2 focuses exclusively on updating the initialized weights $E'$ and $H'$, resulting in a rapid decrease in loss. However, the primary objective of embedding and LM Head training is to optimize the entire vocabulary embedding representations. The strategy of updating only a subset of tokens in Ex2 may neglect the relationships, correlations, and representations of tokens within their contextual surroundings, potentially leading to a local optimum. Consequently, it is generally recommended to train all components $(E, E', H, H')$ from the beginning. Additional metrics were assessed using the KoBEST to corroborate these findings. As depicted in Figure~\ref{fig:train_comparision_pt_metrics_kobest_trials}, the metrics for Ex1 stabilize at higher values.

\textit{Memory and computation demand}

In this context, Ex2 demonstrates optimization compared to Ex1, but not a significant difference. Specifically, during the initial training phase in Ex2, the number of parameters requiring updates is around 140 million (1.29\% trainable parameters) due to the newly initialized weights $(E', H')$. This results in lower memory usage and decreased computational demand. In contrast, Ex1 consistently maintains around 370 million parameters (3.41\% trainable parameters) throughout the entire process. However, it is important to note that this initial training phase in Ex2 is relatively brief, lasting only up to 1,440 steps, and the model was not improved further. Additionally, the subsequent increase in the number of trainable parameters from 140 million to 370 million is not significantly large within the context of LLMs. The trainable parameters of Ex1 and Ex2 in Embedding and LM Head training are shown in Figure~\ref{fig:train_comparision_pt_params_trials}.

In summary, the experimental results, which meticulously consider model performance alongside memory and computational demands, provide evidence that training the entire Embedding and LM Head from the beginning is the preferable approach in our research. Thus, the complex techniques introduced in EEVE may be deemed unnecessary for certain practical continual pretraining processes.

\subsection{Main Results}
\paragraph{Cost}
The training process was conducted using a single Nvidia H100 GPU, with an approximate training duration of 498 GPU hours. The H100 GPU offers a theoretical BF16 training performance of 989 TFLOPS, excluding sparsity. Assuming a Model FLOPS Utilization (MFU) of 40\% and equivalence between Model FLOPS Utilization (MFU) and Hardware FLOPS Utilization (HFU) (without considering activation recomputation), the total computational cost for model training is estimated to be approximately 700 ZFLOPS. Due to the division of the training process into four distinct stages (as detailed in subsection~\ref{subsec:training}) and the complexity of the experimental procedure, an in-depth analysis of training time was not conducted.

Nevertheless, the findings of this research indicate the potential for substantial reductions in training time through the application of advanced continual pretraining and our proposed enhancements. Firstly, data filtering effectively reduces the size of the training dataset while preserving its generalizability. Secondly, the development of an efficient tokenizer tailored for the Korean language significantly decreases input complexity, thereby reducing the computational resources required for training. Thirdly, optimal model initialization provides a more favorable starting point, closer to the target optimization, thus lowering training costs. Finally, the step-by-step training strategy mitigates memory constraints and optimizes training costs while maintaining overall performance. Each component of this carefully designed pipeline contributes to the construction of high-quality models in a cost-effective manner, facilitating more efficient and scalable model training processes.

\paragraph{Results on Pre-trained Models}



\begin{figure}[ht]
\centering
\begin{minipage}[t]{0.45\textwidth}
\centering
\includegraphics[width=\linewidth]{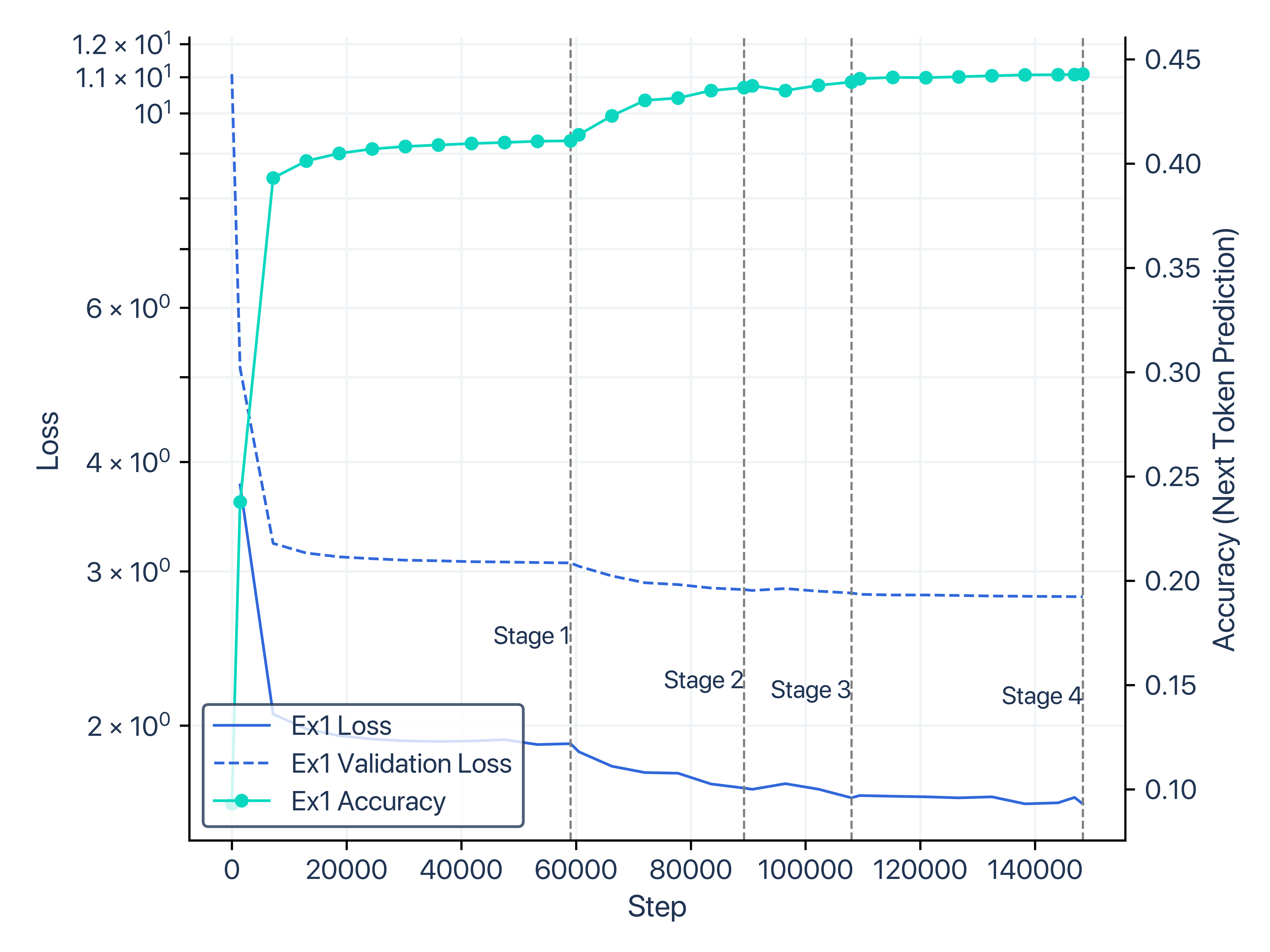}
\caption{Training and evaluation loss.}
\label{fig:train_loss}
\end{minipage}
\hfill
\begin{minipage}[t]{0.45\textwidth}
\centering
\includegraphics[width=\linewidth]{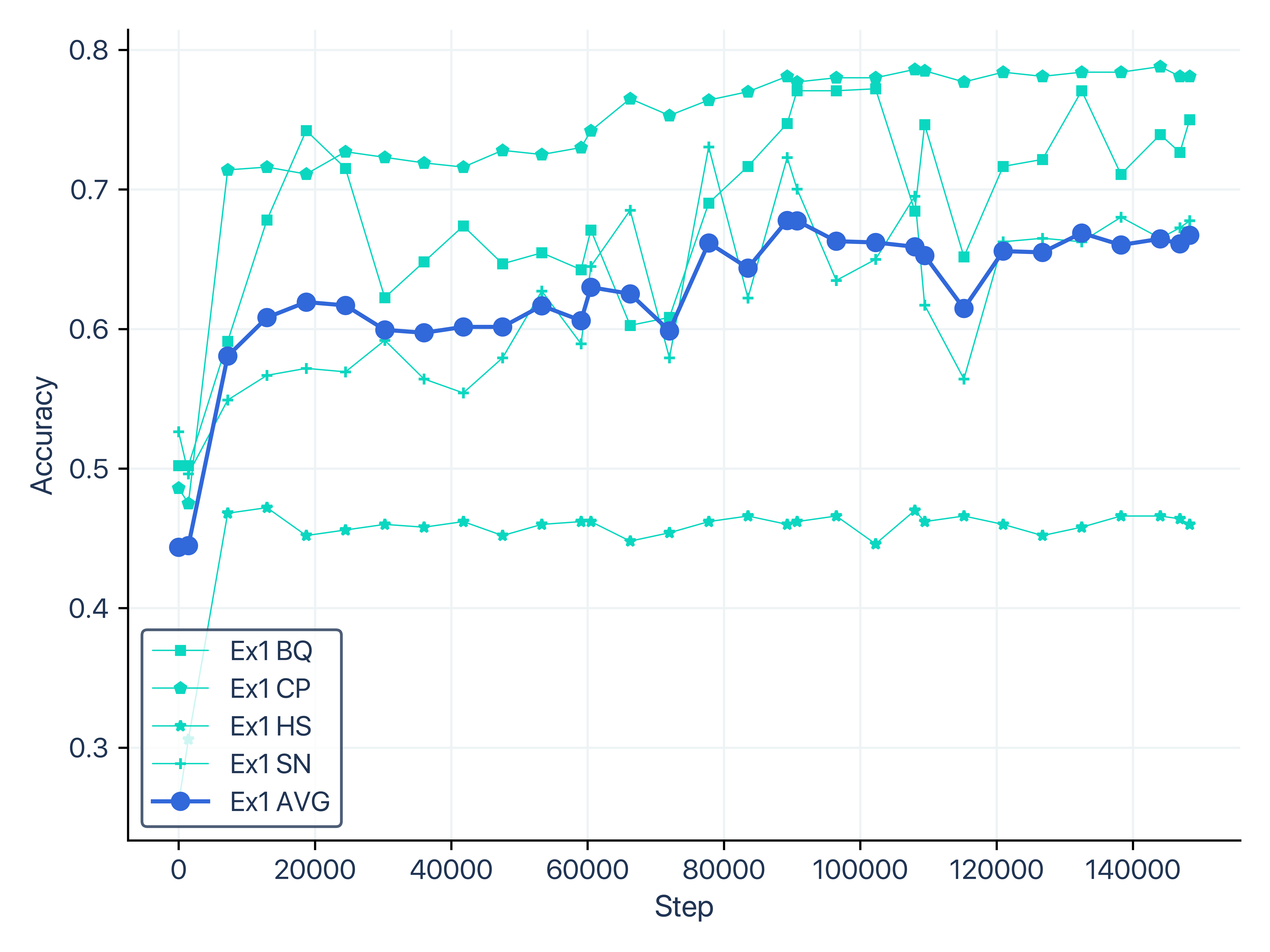}
\caption{KoBEST evaluation.}
\label{fig:train_kobest}
\end{minipage}
\end{figure}

Figure ~\ref{fig:train_loss} shows the training and evaluation loss throughout the entire training process based on Ex1. The solid blue line represents the training loss, which steadily decreases over time. The dashed blue line represents the evaluation loss, which also consistently trends downward. Additionally, the gray line representing the accuracy metric shows an upward trend, indicating improvements in the model's performance on the next token prediction task. These trends suggest that the model is learning effectively in the causal language modeling task.

Figure ~\ref{fig:train_kobest} presents the detailed KoBEST metrics over the training steps. The average accuracy (Ex1 AVG), depicted by the bold blue line, shows an overall upward trend, indicating improved model performance in KoBEST. The performance on specific tasks, such as BoolQ, COPA, HellaSwag, and SentiNeg, also demonstrates variability and improvement throughout the training process. These results highlight the model's ability to generalize and perform well across different tasks that require advanced Korean linguistic knowledge.

Furthermore, it was observed that even after pre-training on 9.7 billion tokens, our model did not exhibit any signs of convergence, indicating potential for further improvement with additional training.

\paragraph{Comparison on General-Purpose Tasks}

\begin{table}[htbp]
\centering
\caption{Comparison of model performance}
\label{table:model-performance-comparision}

\resizebox{\textwidth}{!}{

\begin{tabular}{llrrrrrrrr}
\toprule
\multicolumn{2}{l}{\textbf{Model}} & \multicolumn{3}{l}{\textbf{CLM Metrics}} & \multicolumn{5}{l}{\textbf{KoBEST}} \\
\cmidrule(r){1-2} \cmidrule(r){3-5} \cmidrule(r){6-10}
\textbf{Name} & \textbf{Type} & \textbf{Loss} & \textbf{Eval Loss} & \textbf{Eval Accuracy} & \textbf{BQ} & \textbf{CP} & \textbf{HS} & \textbf{SN} & \textbf{AVG} \\
\midrule
meta-llama/Llama-2-7b-hf & PT & - & 1.2514 & 0.6549	& 0.5199 & 0.5710 &	0.4260	& 0.5919 & 0.5272 \\
meta-llama/Llama-2-7b-chat-hf & FT & - & 1.6147	& 0.6251 & 0.6446 & 0.5550 & 0.4020	& 0.5164 & 0.5295 \\

meta-llama/Llama-2-13b-hf & PT & - & 1.1855	& 0.6712	& 0.5783	& 0.5960	& 0.4240	& 0.5617 & 0.5400 \\
meta-llama/Llama-2-13b-chat-hf & FT & - & 1.4888 & 0.6404 & 0.7614 & 0.574 & 0.396 & 0.5214 & 0.5632 \\

mistralai/Mistral-7B-v0.1 & PT & - & 1.5937	& 0.6117	& 0.7051	& 0.5930	& 0.4380	& 0.5793	& 0.5789 \\
mistralai/Mistral-7B-v0.1-chat & FT & - & 2.0644	& 0.5303 & 0.6531 & 0.5310 & 0.3960 & 0.4962 & 0.5191 \\

microsoft/Phi-2	& PT & - & 2.3143	& 0.4036	& 0.5036	& 0.4820	& 0.328	& 0.4912	& 0.4512 \\
daekeun-ml/phi-2-ko-v0.1 & PT & - & 4.9800 & 0.2341	& 0.5783	& 0.609	& 0.378	& 0.6524	& 0.5544 \\

yanolja/EEVE-Korean-2.8B-v1.0	& PT & - & 5.1550	& 0.2351	& 0.5221	& 0.5870	& 0.372	& 0.3689	& 0.4625 \\
yanolja/EEVE-Korean-Instruct-2.8B-v1.0	& FT & - & 0.2293 & 5.2079	& 0.7386	& 0.5760	& 0.362	& 0.5038	& 0.5451 \\

beomi/OPEN-SOLAR-KO-10.7B	& PT & - & 2.7548	& 0.4484	& 0.8533	& 0.805	& 0.504	& 0.5113	& 0.6684 \\
beomi/open-llama-2-ko-7b & PT & - & 2.9543	& 0.4234	& 0.5014	& 0.7670	& 0.4820	& 0.5113	& 0.5654 \\
beomi/llama-2-koen-13b	& PT & - & 2.8017	& 0.4366	& 0.7556	& 0.7950	& 0.4980	& 0.5365	& 0.6463 \\


upstage/SOLAR-10.7B-v1.0	& PT & - & 2.0461	& 0.5273	& 0.5021	& 0.5790	& 0.4320	& 0.4962	& 0.5023 \\
upstage/SOLAR-10.7B-Instruct-v1.0	& FT & - & 1.9142 & 0.5540 & 0.8860 & 0.6500 & 0.4620 & 0.4937 & 0.6229 \\

yanolja/EEVE-Korean-10.8B-v1.0	& PT & - & 2.9533 & 0.4201 & 0.6880	& 0.751	& 0.4900 & 0.7254 & 0.6636 \\
yanolja/EEVE-Korean-Instruct-10.8B-v1.0	&  FT & - & 2.7436	& 0.4471	& 0.8554	& 0.7510	& 0.4900	& 0.8514	& 0.7370 \\

\midrule
RedWhale-Initilization (our) & PT & - & 11.0905 & 0.093 & 0.5021 & 0.4860	& 0.2600 & 0.5264 & 0.4436 \\
RedWhale-Emb \& Head (our)& PT & 1.9072 & 3.0672 & 0.4109	& 0.6425 & 0.7300 & 0.4620 & 0.5894	& 0.6060 \\
RedWhale-Odd Layers (our)& PT & 1.6970 & 2.8595 & 0.4364	& 0.7472 & 0.7810 & 0.4600 & 0.7229	& 0.6778 \\
RedWhale-Even Layers (our)& PT & 1.6541 & 2.8341 & 0.4391	& 0.6845 & 0.7860 & 0.4700 & 0.6952	& 0.6589 \\
RedWhale-LoRA (our)& PT & 1.6279	& 2.8064 & 0.4429 & 0.7500	& 0.7810	& 0.4600	& 0.6776	& 0.6672 \\
\midrule
yanolja/EEVE-Korean-10.8B-v1.0-SFT & FT & 1.2050 &  2.7100 &  0.4512 & 0.8632  & 0.8840  &  0.4600 	&  0.9698 	& 0.7942  \\
\textbf{RedWhale-SFT} (our)& FT & 1.2288 &  2.8752 & 0.4330 & 0.9274 & 0.8830 &  0.4580 &  0.9647 &  \textbf{0.8083} \\

\bottomrule
\end{tabular}
}
\end{table}

\begin{figure}[ht]
\centering
\begin{minipage}[t]{0.48\textwidth}
\centering
\includegraphics[width=\linewidth]{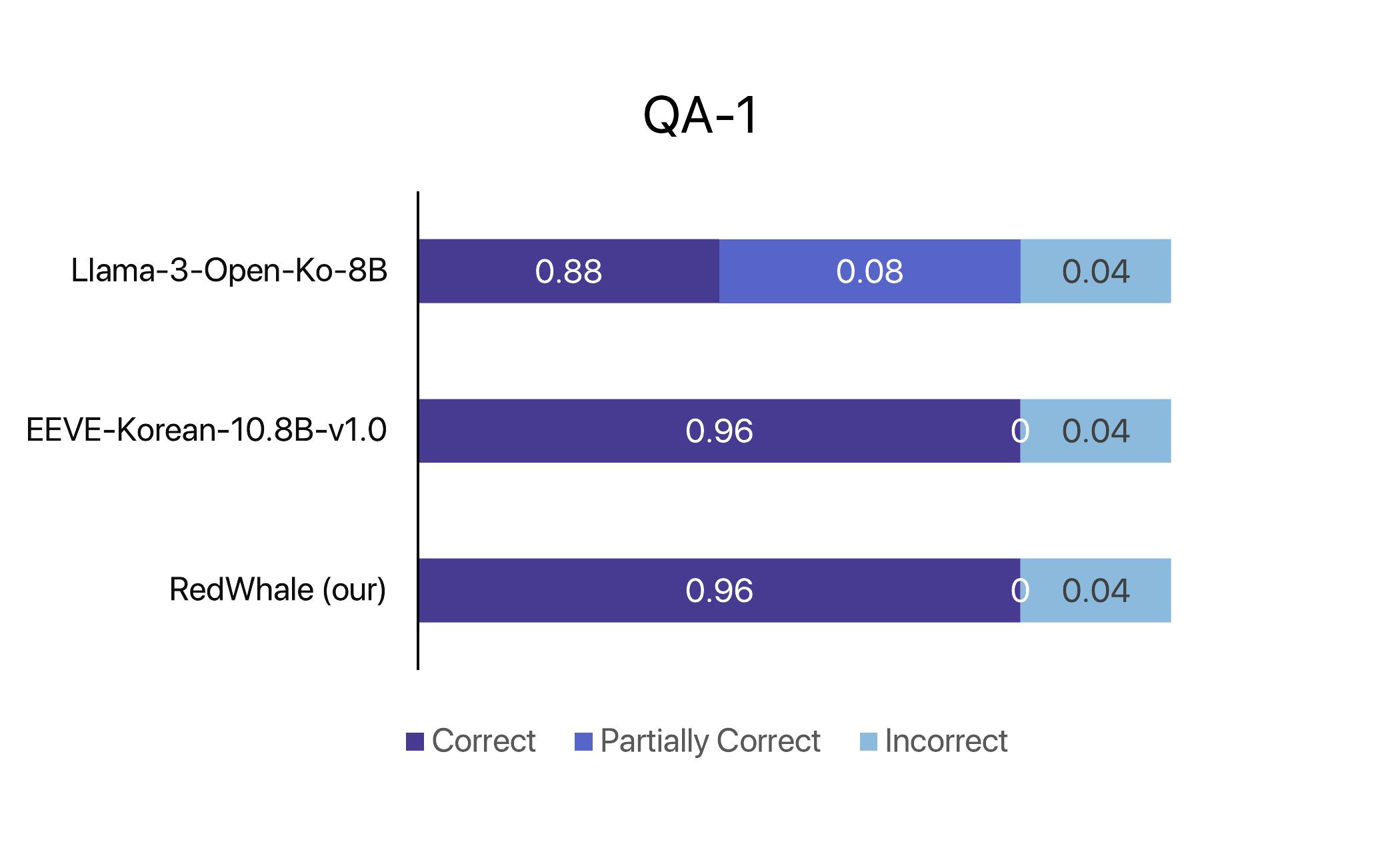}
\caption{Financial QA-1 evaluation.}
\label{fig:fig_financial_rag_1}
\end{minipage}
\hfill
\begin{minipage}[t]{0.48\textwidth}
\centering
\includegraphics[width=\linewidth]{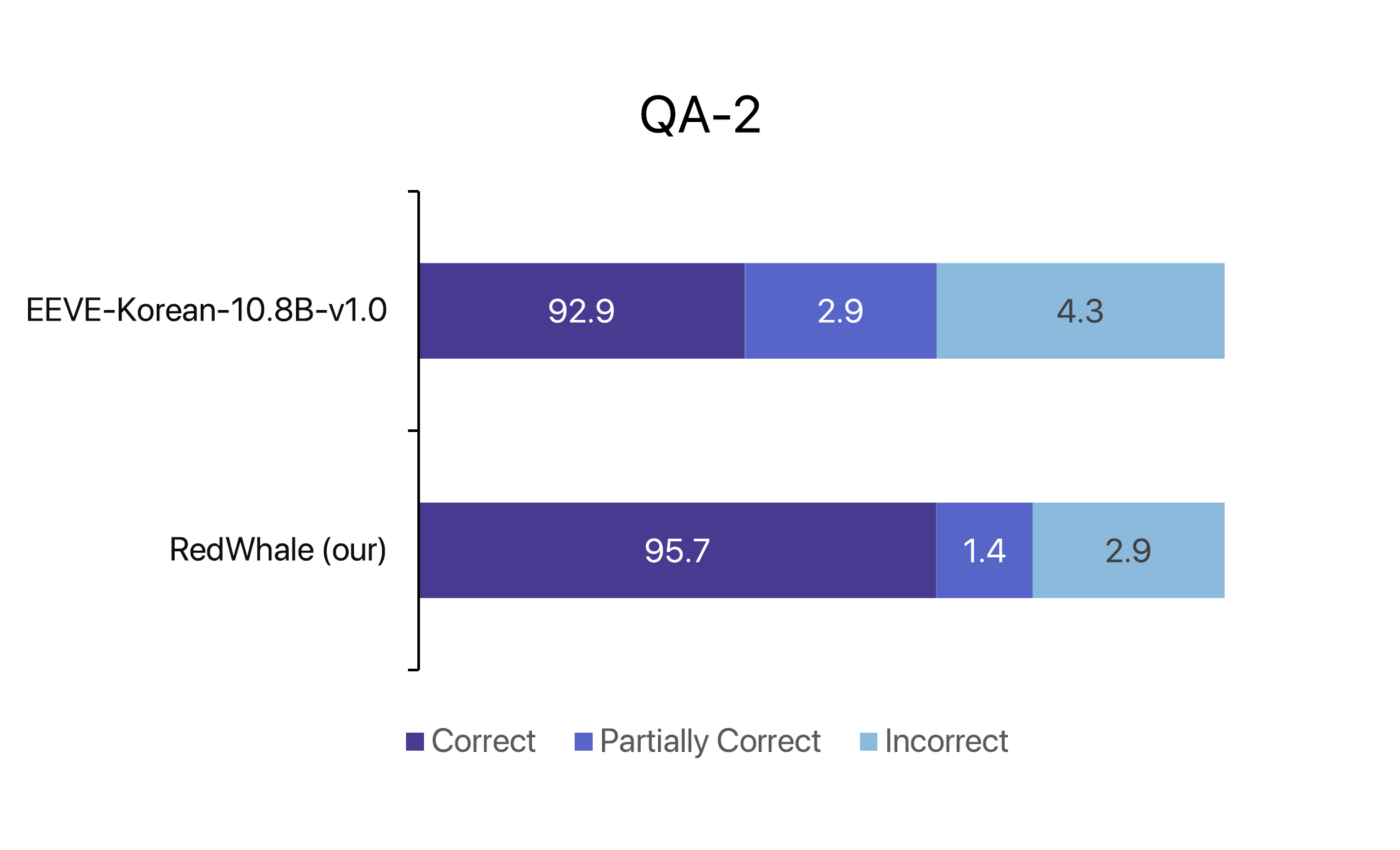}
\caption{Financial QA-2 evaluation.}
\label{fig:fig_financial_rag_2}
\end{minipage}
\end{figure}

Table \ref{table:model-performance-comparision} presents a comprehensive comparison of our model with various reference models.

The CLM metrics, including loss and next-token prediction accuracy measurements, are primarily used to monitor the learning process of a model and are not intended for comparison between different models due to differences in token vocabulary.

In contrast, KoBEST metrics can be used to compare models on specific tasks. These tasks, designed by professional Korean linguists, indicate the model's quality for tasks requiring advanced Korean linguistic knowledge.

The experimental results demonstrate that our model outperforms other top-performing models, including EEVE (\cite{kim2024efficient}). Our model shows a slight improvement, with gains of 0.36\% and 1.41\% for the pretrained and fine-tuned models, respectively. Specifically, we achieved substantial improvements over the SOLAR base model (\cite{kim2023solar}), increasing the KoBEST score from 50.23\% to 66.72\% and from 62.29\% to 80.83\% for the pretrained and fine-tuned models, respectively. This demonstrates our superior Korean processing capabilities in adapting an English-centric model to Korean.

\paragraph{Comparison on Financial Tasks} 

Figures~\ref{fig:fig_financial_rag_1} and~\ref{fig:fig_financial_rag_2} present the experimental results of the Korean Financial QA tasks, comparing our model with the reference models. The Llama3 model\footnote{\url{https://huggingface.co/beomi/Llama-3-Open-Ko-8B}} represents an early effort to adapt Llama3 for the Korean language. In contrast, our model and the EEVE model are more advanced attempts to tailor the model for Korean, aiming for high performance in an efficient manner. The experimental outcomes indicate that our model performs equivalently to the EEVE model on QA-1 tasks, but demonstrates slightly better performance on QA-2 tasks. Both models outperform the Llama3 model.

These findings underscore three key points: First, they emphasize the significance of language-specific adaptations in enhancing model performance, as demonstrated by the superior outcomes of both our model and the EEVE model in comparison to the Llama3 model. Second, they reveal nuanced differences in task performance, particularly in QA-2 tasks, where our model's marginal advantage suggests that continual pretraining optimization significantly facilitates downstream applications, yielding notable improvements, especially in complex financial queries. Finally, the experimental results indicate the necessity for general models to undergo an additional phase of domain-specific adaptation, particularly in specialized fields such as finance.

\section{Discussion}
The methodology proposed in this study is categorized under ``Continual Pretraining'' or ``Further Pretraining,'' particularly within the context of language adaptation. This approach serves as a vital intermediary stage between the initial pretraining phase and subsequent domain-specific adaptation or alignment. It enables the efficient adaptation of English-centric language models to new languages, such as Korean, or specific domains, all while requiring relatively minimal learning and computational resources.

\textbf{Feasibility with Limited Learning and Computational Resources}

Our approach demonstrates that language adaptation can be achieved with a reduced amount of learning and limited computational resources due to several key factors:

\begin{itemize}
    \item \textbf{Corpus filtering:} By employing a data filtering strategy, we significantly reduce the dataset size, which in turn decreases the number of learning steps required.
    \item \textbf{Tokenizer Adaptation:} Adjusting the tokenizer to handle shorter input lengths further reduces the number of learning steps, facilitating quicker training.
    \item \textbf{Model initialization:} Starting from an optimal model initialization point ensures fewer learning steps are needed to achieve convergence.
    \item \textbf{Model training strategy:} The staged-training strategy, which incorporates transfer learning, significantly reduces the number of learning steps. This is achieved by training new model components with a large learning rate for rapid adaptation, while tuning older components with a smaller learning rate. Additionally, the use of parameter-efficient LoRA techniques consolidates learning across all components, maintaining computational efficiency.
    \item \textbf{Limited computational resources:} The staged-training strategy also addresses memory constraints by training different parts of the model sequentially. This approach prevents out-of-memory errors, with each training step designed to require minimal memory and be completed within a reasonable timeframe, making the methodology practical and feasible even with limited hardware resources.
\end{itemize}

\textbf{Method Efficiency}

The efficiency of our method, especially when compared to standard training approaches, is highlighted by its cost-effectiveness and hardware flexibility. This approach offers substantial savings in GPU hours and overall training costs. Moreover, it enables training on less powerful hardware, such as A6000 48GB or multiple A5000 24GB GPUs, rather than relying on more expensive and resource-intensive GPUs like the H100 80GB.

\textbf{Model Efficiency}

\begin{figure}[ht]
\centering
\includegraphics[width=0.8\linewidth]{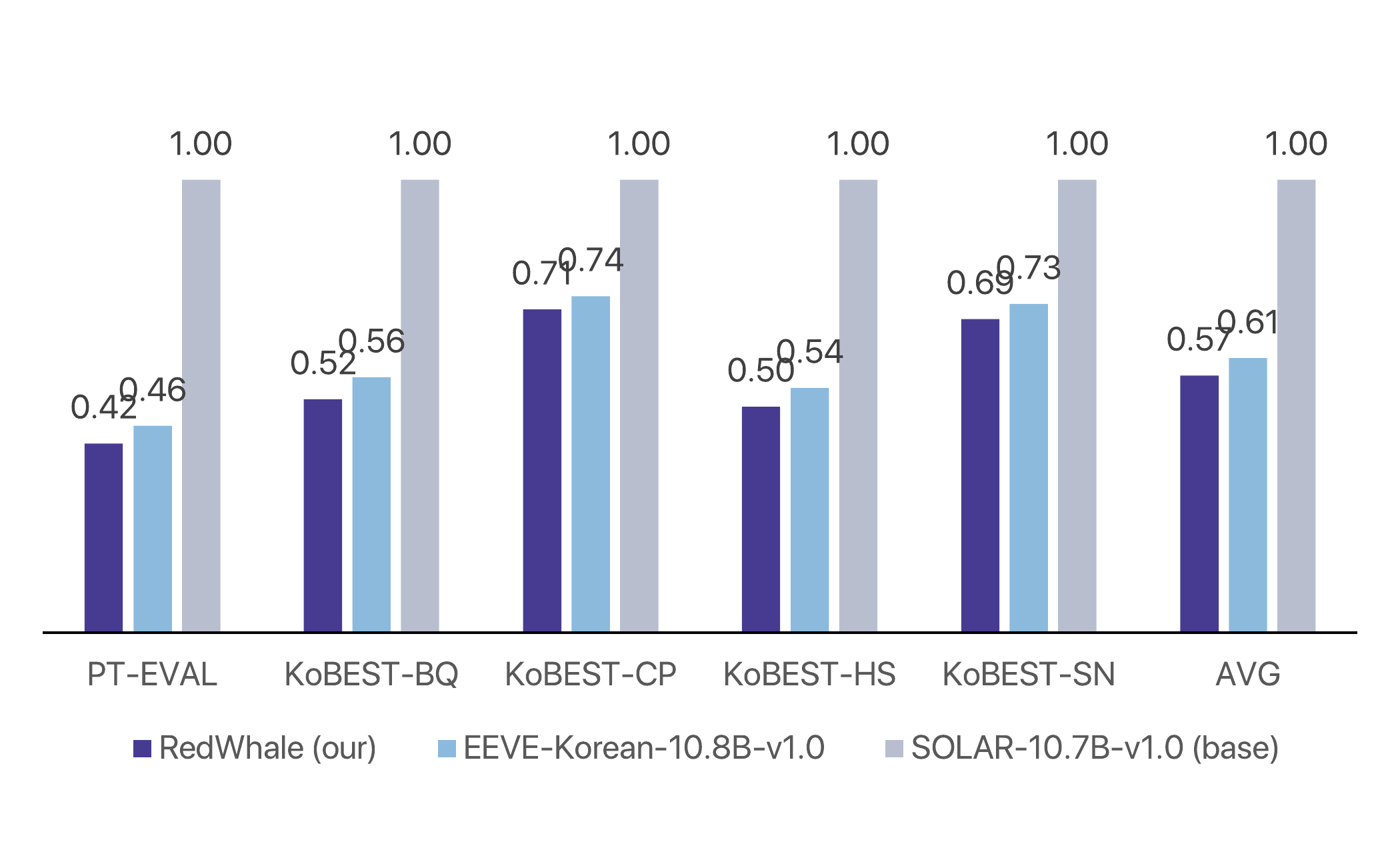}
\caption{Token Ratio (TR).}
\label{fig:fig_token_count_ratio}
\end{figure}

The resulting model from this experiment, named RedWhale, demonstrates significant efficiency improvements. RedWhale achieves the following:

\begin{enumerate}
    \item \textbf{Faster processing and improved handling of longer documents:} This is largely due to the adaptation of the tokenizer to the Korean language.
    \item \textbf{Reduced fine-tuning costs for downstream tasks:} The model provides an optimal starting point, lowering the costs associated with fine-tuning.
    \item \textbf{Further customization with minimal additional expense:} The model allows for additional customization without incurring substantial costs.
\end{enumerate}

In addition to the model quality evidenced by the KoBEST benchmark and financial tasks, the Token Ratio (TR) serves as an indirect measure of the observed efficiency improvements. The TR is determined by dividing the number of tokens produced by the newly developed tokenizer by the number of tokens generated by the base tokenizer when applied to the same dataset, thereby providing a normalized metric of token generation efficiency. A lower TR is indicative of enhanced effectiveness, as it reflects the new tokenizer's ability to produce fewer tokens than the base model for identical input data. This reduction in token count offers three principal advantages: (1) a linear decrease in the number of inputs to the model, thereby reducing the required learning steps; (2) the ability to process longer input texts within the fixed context length of the model; and (3) faster processing of input data due to the reduced number of tokens. In this study, the tokenizer was adapted for the Korean language, resulting in an average TR of 0.57 across five datasets, including the pretraining evaluation set and four related tasks on KoBEST, as depicted in Figure \ref{fig:fig_token_count_ratio}. This adaptation led to a 50\% reduction in token count, the capability to handle sequences over four times longer on the same hardware—considering the quadratic increase in computational complexity of Transformers with respect to token length—and, consequently, faster training and inference relative to the base model.

\textbf{Use Case Advantages}

This methodology excels in scenarios where hardware limitations or budget constraints are significant. It offers a practical solution for those without access to powerful hardware, enabling training that would otherwise be infeasible due to out-of-memory errors or excessively long training times. Conversely, if powerful hardware is available and budget is not a concern, direct training from scratch or continual pretraining using standard methods like LLaMA (\cite{touvron2023llama}) or Mistral (\cite{jiang2023mistral}) may be more feasible and straightforward.

\textbf{Methodological Innovations}

Although this method is not entirely novel, it represents a strategic enhancement and combination of existing approaches, introducing several key improvements:

\begin{itemize}
    \item \textbf{Corpus processing:} While this step builds on established methodologies, we have carefully analyzed and evaluated its application to ensure optimal outcomes.
    \item \textbf{Tokenizer:} We have refined existing tokenizer adaptation techniques by removing unnecessary tokens from the vocabulary and conducting a thorough analysis to determine the optimal vocabulary size. These improvements contribute to a more efficient and accurate tokenization process.
    \item \textbf{Initialization:} We extended the application of existing initialization methods beyond the input embeddings to include the output (unembedding) components as well. Multiple methods were tested to identify the most effective initialization strategy, ensuring optimal model performance.
    \item \textbf{Training:} The training process integrates elements from both Chinese (\cite{cui2024efficient}) and Korean (\cite{kim2024efficient}) model adaptation methods. We streamlined the Korean approach by removing inefficient steps to reduce complexity and cost, while incorporating enhancements from the Chinese method to boost performance. Notably, our strategy of quickly adapting unpretrained modules, followed by training 50\% of the model at a time—focusing on layers in odd and even positions—before integrating all components with LoRA, represents a novel approach not commonly referenced in existing literature.
\end{itemize}

\textbf{Limitations and Challenges}

\textit{Corpus Filtering Challenges:} This study employs corpus filtering with the hypothesis that careful selection and refinement of data can enhance the consistency and quality of each sample, potentially leading to improved computational efficiency and model performance. Although a qualitative analysis through sampling has been conducted to ensure optimal outcomes, a formal quantitative evaluation of the final results remains to be completed. Two primary concerns arise from this process: first, the possibility that utilizing "high-quality" data sources may not lead to improved performance or sustain diversity coverage distribution when compared to random data selection; and second, the risk that improper filtering methods could degrade model quality. For instance, perplexity-based filtering, which typically offers high precision but low recall, may inadvertently exclude numerous clear and relevant documents despite the high quality of those selected.

In addition to the methods already applied, it is essential that data filtering procedures consider, evaluate, and compare various approaches. These approaches include random filtering (to achieve high generalizability), filtering based on text length (assuming that longer documents may be superior to shorter or medium-length ones), perplexity-based filtering (examining low, moderate, and high perplexity levels), filtering based on diversity (using metrics such as Type-Token Ratio (TTR) or clustering), and assessing the impact of known and unknown samples on model performance. Moreover, the pretraining process, which theoretically benefits from learning across large datasets that may include noise, raises critical questions regarding the necessity and impact of filtering. To address these challenges, future research will focus on two key questions: (1) Does filtering genuinely enhance model performance by reducing the number of training samples while maintaining accuracy? and (2) If so, what is the most effective and appropriate filtering method?

\textit{Staged Training Strategy Limitations:} The three-stage training strategy employed in this study is structured as follows: (1) Initially, new modules (Embedding and LM Head) are trained with a high learning rate to facilitate rapid adaptation; (2) Subsequently, existing components (Transformer blocks) are fine-tuned with a lower learning rate; and (3) Finally, all components undergo training using LoRA (a parameter-efficient approach) to consolidate learning. Each stage can be further subdivided into specific sub-stages. For instance, during the training of Transformer blocks, layers in odd-numbered positions (e.g., 1, 3, 5) are trained first, followed by layers in even-numbered positions (e.g., 0, 2, 4). This approach is hypothesized to offer two key advantages: it is feasible with limited GPU memory, and it allows different parts of the model to learn distinct features of the data. However, due to the absence of comparative experiments, such as training layers sequentially from the lower (0, L/2) to the upper half (L/2, L) of the model (where L represents the total number of layers), the efficacy of this strategy remains unvalidated. Consequently, the method employed may not represent the optimal solution.

\textit{Performance Evaluation Limitations:} The lack of comparable research on similar experiments has precluded a detailed cost-effectiveness comparison. Although model performance evaluations on KoBEST and Financial QA demonstrated superior performance in these specific tasks, the broader challenge of evaluating a large language model persists. Further evaluations are necessary to confirm the superiority of the proposed model and methodology.

\section{Conclusion}

In this study, we present RedWhale, the output model of a proposed efficient continual pretraining method specifically designed for adapting English LLMs to Korean. Addressing the distinctive challenges of Korean's non-alphabetic token structure and the intensive computational requirements of large language model training, our approach integrates several key contributions. Our preprocessing pipeline significantly improved the quality of the Korean corpus, ensuring high-quality data and reducing computational demands. We developed an advanced tokenizer tailored to Korean, optimizing the balance between input and embedding complexity, resulting in a more efficient tokenizer. Through optimal initialization methods, we minimized training expenses and computational efforts, particularly for the Embedding and Language Model Head components. Our four-stage training strategy accommodated hardware limitations and ensured thorough pretraining, while the application of LoRA and instruction tuning further refined the model's alignment and task-specific performance. Experimental results show that RedWhale outperforms other leading models on Korean NLP benchmarks, such as KoBEST. Notably, our model showed no signs of convergence even after pretraining on 9.7 billion tokens, indicating potential for further enhancement with additional training. RedWhale represents a significant advancement in Korean language processing capabilities, addressing specific challenges and offering insights for building high-quality models in a cost-effective manner for other low-resource languages or domain-specific adaptations.

\section{Acknowledgment}
This work was fully supported by AGILESODA Inc.

\bibliographystyle{unsrtnat}
\bibliography{references}  






\end{document}